%% file: main.tex
\documentclass{article} 
\usepackage{iclr2022_conference,times}

\input{macros.tex}

\title{NAS-Bench-Suite: \\ NAS Evaluation is (Now) Surprisingly Easy}

\author{Yash Mehta$^1$\thanks{Equal contribution.

Email to:
\texttt{\{mehtay,zelaa\}@cs.uni-freiburg.de}, \texttt{colin@abacus.ai}}, ~Colin White$^2$\footnotemark[1], ~
Arber Zela$^1$, ~Arjun Krishnakumar$^1$, \\
    \textbf{Guri Zabergja$^1$, Shakiba Moradian$^1$, Mahmoud Safari$^1$, Kaicheng Yu$^2$, Frank Hutter$^{1,3}$}
    \vspace*{1mm}\\
    $^1$ University of Freiburg, $^2$ Abacus.AI, $^3$ Bosch Center for AI
}

\iclrfinalcopy
\begin{document}
\maketitle

\input{sections/0_abstract}
\input{sections/1_introduction}

\input{sections/3_benchmark_overview}

\input{sections/4_experiments}

\input{sections/5_codebase}

\input{sections/8_related_work}

\input{sections/9_conclusion}


\input{sections/10_ethics_statement}

\subsubsection*{Acknowledgments and Disclosure of Funding}
FH and his group acknowledge support by the German Federal Ministry of Education and Research (BMBF, grant RenormalizedFlows 01IS19077C and grant DeToL), the Robert Bosch GmbH, the European Research Council (ERC) under the European Union Horizon 2020 research and innovation programme through grant no. 716721, and by TAILOR, a project funded by EU Horizon 2020 research and innovation programme under GA No 952215. This research was funded by the Deutsche Forschungsgemeinschaft (DFG, German Research Foundation) under grant number 417962828. 
We thank Danny Stoll and Falak Vora for their helpful contributions to this project.

\bibliography{main}
\bibliographystyle{iclr2022_conference}

\newpage
\appendix

\input{sections/11a_nas_reproducibility}

\input{sections/A_appendix_benchmarks}

\input{sections/B_appendix_experiments}

\input{sections/C_appendix_codebase}

\end{document}

%% file: macros.tex
\usepackage{amsmath,amsfonts,bm}

\usepackage[utf8]{inputenc} 
\usepackage[T1]{fontenc}    
\usepackage{hyperref}       
\usepackage{url}            
\usepackage{booktabs}       
\usepackage{amsfonts}       
\usepackage{nicefrac}       
\usepackage{microtype}      
\usepackage{xcolor}         

\usepackage{comment}
\usepackage{pifont}
\usepackage[nointegrals]{wasysym}

\usepackage[pdftex]{graphicx}
\usepackage{amssymb}
\usepackage{multirow}
\usepackage{xspace}
\usepackage{adjustbox}
\usepackage{algorithm2e,algorithmic}
\usepackage{listings}
\usepackage{subcaption}
\usepackage{enumitem}
\usepackage{wrapfig}

\def\eqref#1{equation~\ref{#1}}

\def\1{\bm{1}}

\DeclareMathAlphabet{\mathsfit}{\encodingdefault}{\sfdefault}{m}{sl}
\SetMathAlphabet{\mathsfit}{bold}{\encodingdefault}{\sfdefault}{bx}{n}

\definecolor{codegreen}{rgb}{0,0.6,0}
\definecolor{codegray}{rgb}{0.5,0.5,0.5}
\definecolor{codepurple}{rgb}{0.58,0,0.82}
\definecolor{backcolour}{HTML}{F2F2F2}

\usepackage[]{todonotes}

\newcommand{\cmark}{\ding{51}}%
\newcommand{\nasbs}{\texttt{NAS-Bench-Suite}}%

\newcommand{\answerTODO}[1][]{\textcolor{red}{\bf [TODO]}}

%% file: sections/0_abstract.tex
\begin{abstract}

The release of tabular benchmarks, such as NAS-Bench-101 and NAS-Bench-201, has significantly lowered the computational overhead for conducting scientific research in neural architecture search (NAS). Although they have been widely adopted and used to tune real-world NAS algorithms, these benchmarks are limited to small search spaces and focus solely on image classification. Recently, several new NAS benchmarks have been introduced that cover significantly larger search spaces over a wide range of tasks, including object detection, speech recognition, and natural language processing. However, substantial differences among these NAS benchmarks have so far prevented their widespread adoption, limiting researchers to using just a few benchmarks. In this work, we present an in-depth analysis of popular NAS algorithms and performance prediction methods across 25 different combinations of search spaces and datasets, finding that many conclusions drawn from a few NAS benchmarks do \emph{not} generalize to other benchmarks. To help remedy this problem, we introduce \nasbs, a comprehensive and extensible collection of NAS benchmarks, accessible through a unified interface, created with the aim to facilitate reproducible, generalizable, and rapid NAS research. Our code is available at \url{https://github.com/automl/naslib}.

\end{abstract}

%% file: sections/1_introduction.tex
\section{Introduction}
\label{sec:intro}

Automated methods for neural network design, referred to as neural architecture search (NAS), have been used to find architectures that are more efficient and more accurate than the best manually designed architectures~\citep{zoph2018learning, real2019regularized, so2019evolved}. However, it is notoriously challenging to provide fair comparisons among NAS methods due to potentially high computational complexity~\citep{zoph2017neural, real2019regularized} and the use of different training pipelines and search spaces~\citep{randomnas, lindauer2019best}, resulting in the conclusion that \emph{``NAS evaluation is frustratingly hard''}~\citep{yang2019evaluation}. To make fair, statistically sound comparisons of NAS methods more accessible, tabular NAS benchmarks have been released; these exhaustively evaluate all architectures in a given search space, storing the relevant training metrics in a lookup table~\citep{nasbench, nasbench201, nasbench1shot1, nasbenchasr}. This substantially lowers the computational overhead of NAS experiments, since the performance of an architecture can be found simply by querying these tables, hence allowing for a rigorous comparison of various NAS algorithms with minimal computation.

While early tabular NAS benchmarks, such as NAS-Bench-101~\citep{nasbench} and NAS-Bench-201 \citep{nasbench201}, have been widely adopted by the community, they are limited to small search spaces and focus solely on image classification. Recently, benchmarks have been introduced for natural language processing~\citep{nasbenchnlp}, speech recognition~\citep{nasbenchasr}, object detection, and self-supervised tasks~\citep{transnasbench}. Furthermore, the release of \emph{surrogate} NAS benchmarks~\citep{nasbench301, nasbenchx11}, which estimate the performance of all architectures in a search space via a surrogate model, has removed the constraint of exhaustively evaluating the entire search space, expanding the scope of possible search space sizes to $10^{18}$ and beyond. 
However, substantial differences in the abstractions (such as whether a node or an edge denotes an operation), capabilities (such as whether all, or only some, of the architectures can be queried), and implementations (such as incompatible deep learning libraries) have so far prevented nearly all research in NAS from providing results on more than two families of benchmarks. Overall, the lack of consistency in ``NAS-Bench'' datasets has significantly slowed their collective adoption.

In this work, we show that there is a need to adopt newer benchmarks because many conclusions drawn from a small subset of benchmarks do not generalize 
across diverse datasets and tasks.
Specifically, we present an in-depth analysis of popular black-box \citep{real2019regularized,bananas,ottelander2020local}, one-shot \citep{darts,drnas,gdas}, and performance prediction methods \citep{white2021powerful} across (nearly) 
every publicly available queryable NAS benchmark.
This includes 25 different combinations of search spaces and datasets, which is, 
to the best of our knowledge, by far the largest set of NAS search spaces and datasets on which experiments have been conducted to date. 
We show that many implicit assumptions in the NAS community are wrong. 
First, if a NAS algorithm does well on NAS-Bench-101 and NAS-Bench-201, it does not necessarily perform well on other search spaces. 
Second, NAS algorithms may not have robust default hyperparameters and therefore require tuning.
Finally, tuning the hyperparameters of a NAS method on one search space and transferring these
hyperparameters to other search spaces often make the NAS method perform significantly worse.

\begin{figure}[t]
    \centering
    \includegraphics[width=\textwidth]{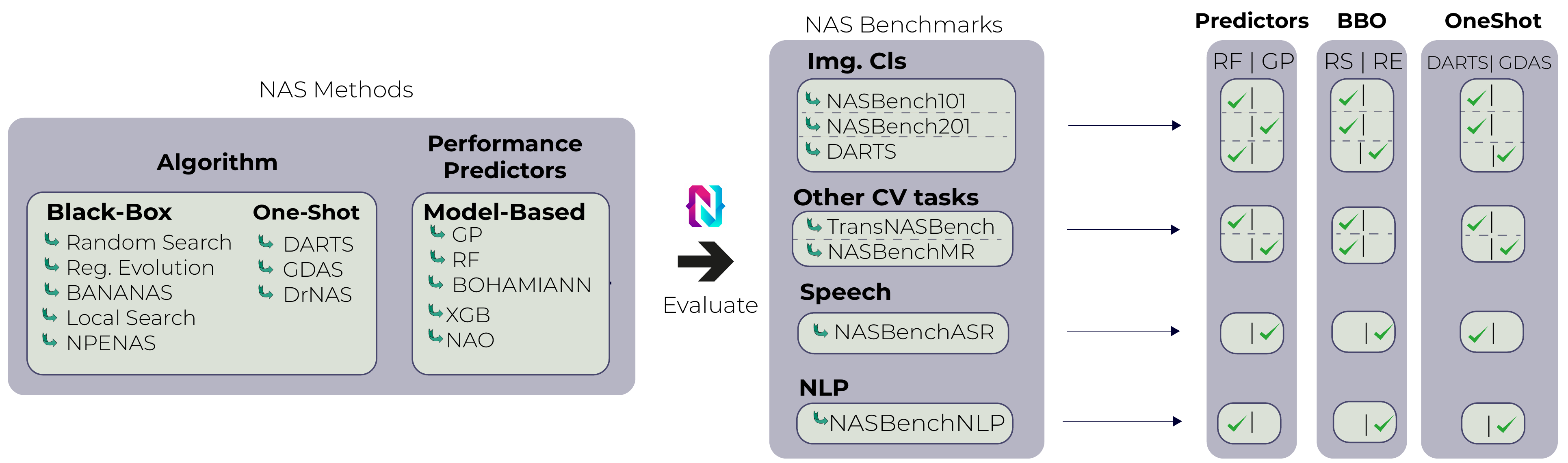}
    \caption{Overview of \nasbs.
    }
    \vspace{-4mm}
    \label{fig:nasbs-overview}
\end{figure}

In order to help NAS researchers and practitioners avoid these pitfalls, we release the NAS Benchmark Suite (\nasbs), a comprehensive and extensible collection of NAS benchmarks, accessible through a unified interface, created with the aim to facilitate reproducible, generalizable, and rapid NAS research. Our work eliminates the overhead for NAS research to evaluate on several different datasets and problem types, helping the community to develop NAS methods that generalize to new problem types and unseen datasets.
See Figure~\ref{fig:nasbs-overview} for an overview.
To ensure reproducibility and other best practices, we release our code and adhere to the NAS best practices checklist~\citep[][see Section~\ref{app:nas_checklist} for details]{lindauer2019best}.

\noindent\textbf{Our contributions.} We summarize our main contributions below.
\begin{itemize}[topsep=0pt, itemsep=2pt, parsep=0pt, leftmargin=5mm]
     \item 
     We conduct a comprehensive study of the generalizability of NAS algorithms and their hyperparameters across 25 settings, showing that it is often not sufficient to tune on just a few benchmarks, and showing that the best hyperparameters depend on the specific search space.
     \item
     We introduce a unified benchmark suite, \nasbs, which implements nearly every publicly available queryable NAS benchmark -- 25 different combinations of search spaces and datasets. By making it easy to quickly and comprehensively evaluate new NAS algorithms on a broad range of problems, our benchmark suite can improve experimental rigor and generalizability in NAS research.
\end{itemize}

%% file: sections/3_benchmark_overview.tex

\vspace{-2mm}
\section{NAS Benchmarks Overview} \label{sec:nasbench_overview}
\vspace{-1mm}


\begin{table}[t]
\caption{Overview of NAS benchmarks in \nasbs{}.}
\centering
\begin{adjustbox}{width=0.9\columnwidth}
\begin{tabular}{@{}lcccccccc@{}}
\multicolumn{1}{l}{} & \multicolumn{1}{c}{} & \multicolumn{2}{c}{\textbf{Queryable}} 
& \multicolumn{1}{c}{} & \multicolumn{1}{c}{} & \multicolumn{1}{c}{} & \multicolumn{1}{c}{} \\
\cmidrule{3-4} \textbf{Benchmark} & \textbf{Size} & \textbf{Tab.} & \textbf{Surr.} 
& \textbf{LCs} & \textbf{Macro} & \textbf{Type} & \textbf{\#Tasks} & \textbf{NAS-Bench-Suite} \\
\midrule 
NAS-Bench-101 & 423k & \cmark & & & & Image\ class.\ & 1 & \cmark \\
\midrule 
NAS-Bench-201 & 6k   & \cmark & & \cmark &  & Image\ class.\ & 3 & \cmark \\
\midrule 
NAS-Bench-NLP & $10^{53}$ & &  & \cmark &  & NLP & 1 & \cmark \\
\midrule 
NAS-Bench-1Shot1 & 364k & \cmark & & & & Image\ class.\ & 1 & \cmark \\
\midrule 
NAS-Bench-301 & $10^{18}$ & & \cmark & & & Image\ class.\ & 1 & \cmark \\
\midrule
NAS-Bench-ASR & 8k & \cmark & & & \cmark & ASR & 1 & \cmark \\
\midrule
NAS-Bench-MR & $10^{23}$ & & \cmark & & \cmark & Var.\ CV & 4 & \cmark \\
\midrule
TransNAS-Bench & 7k & \cmark & & \cmark & \cmark & Var.\ CV & 14 & \cmark \\
\midrule
NAS-Bench-111 & 423k & & \cmark & \cmark & & Image\ class.\ & 1 & \cmark \\
\midrule
NAS-Bench-311 & $10^{18}$ & & \cmark  & \cmark & & Image\ class.\ & 1 & \cmark \\
\midrule
NAS-Bench-NLP11 & $10^{53}$ & & \cmark & \cmark & & NLP & 1 & \cmark \\
\bottomrule
\end{tabular}
\end{adjustbox}
\label{tab:benchmarks}
\end{table}

\paragraph{Preliminaries.}
A \emph{search space} in NAS is the set of all architectures that the NAS algorithm is allowed to select. 
Most recent search spaces are defined by a \emph{cell-based (micro)} structure and a \emph{macro} structure. 
A \emph{cell} is a small set of neural network operations arranged in a
directed acyclic graph (DAG), with constraints on the number of nodes, edges, and
incoming edges per node.
The \emph{macro} structure consists of the architecture skeleton and the 
arrangement of cells, such as how many times each cell is duplicated.
For many popular search spaces, the macro structure is completely fixed,
while for other search spaces, the macro structure can have variable
length, width, and number of channels for different architectures in the search space.

A \emph{NAS benchmark}~\citep{lindauer2019best} consists of a dataset (with a fixed
train-test split), a search space, and a fixed evaluation pipeline with predefined 
hyperparameters for training the architectures.
A \emph{tabular} NAS benchmark is one that additionally provides precomputed evaluations 
with that training pipeline for all possible architectures in the search space.
Finally, a \emph{surrogate} NAS benchmark~\citep{nasbench301, nasbenchx11} 
is a NAS benchmark that provides a surrogate
model that can be used to predict the performance of any architecture in the search
space.
We say that a NAS benchmark is \emph{queryable} if it is either a tabular or surrogate
benchmark.
Queryable NAS benchmarks can be used to simulate NAS experiments very cheaply by querying the performance
of neural networks (using a table or a surrogate) instead of training the neural networks
from scratch.

\noindent\textbf{NAS benchmarks.} 
Now we describe the characteristics of many popular NAS benchmarks.
For a summary, see Table~\ref{tab:benchmarks}, and for a more comprehensive and detailed survey, see Appendix~\ref{sec:appendix_nasbench_overview}.

The first tabular NAS benchmark to be released was NAS-Bench-101~\citep{nasbench}.
This benchmark consists of $423\,624$ architectures trained on CIFAR-10. 
The cell-based search space consists of a directed acyclic graph (DAG) structure in which the nodes can take on operations. A follow-up work, NAS-Bench-1Shot1~\citep{nasbench1shot1}, defines three subsets of NAS-Bench-101 which allow one-shot algorithms to be run. The largest subset size in NAS-Bench-1Shot1 is $363\,648$.
NAS-Bench-201~\cite{nasbench201} is another popular tabular NAS benchmark. The cell-based search space consists of a DAG where each \emph{edge} can take on operations (in contrast to NAS-Bench-101, in which the \emph{nodes} are operations). The number of non-isomorphic architectures is $6\,466$ and all are trained on CIFAR-10, CIFAR-100, and ImageNet-16-120. NATS-Bench~\citep{natsbench} is an extension of NAS-Bench-201 which also varies the macro architecture. 

NAS-Bench-NLP~\citep{nasbenchnlp} is a NAS benchmark for natural language processing, which is size $10^{53}$. However, only $14\,322$ of the architectures were trained on Penn TreeBank~\citep{penntreebank}, meaning NAS-Bench-NLP is not queryable.
NAS-Bench-ASR~\citep{nasbenchasr} is a tabular NAS benchmark for automatic speech recognition. The search space consists of $8\,242$ architectures trained on the TIMIT dataset.
TransNAS-Bench~\citep{transnasbench} is a tabular NAS benchmark consisting of two separate search spaces (cell-level and macro-level) and seven tasks including pixel-level prediction, regression, and self-supervised tasks. The cell and macro search spaces are size $4\,096$ and $3\,256$, respectively.
NAS-Bench-MR \citep{nasbenchmr} is a surrogate NAS benchmark which evaluates across four datasets: ImageNet50-1000, Cityscapes, KITTI, and HMDB51. NAS-Bench-MR consists of a single search space of size $10^{23}$.

The DARTS~\citep{darts} search space with CIFAR-10, consisting of $10^{18}$ architectures, is arguably the most widely-used NAS benchmark. Recently, $60\,000$ of the architectures were trained and used to create NASBench-301~\citep{nasbench301}, the first surrogate NAS benchmark.
More recently, NAS-Bench-111, NAS-Bench-311, and NAS-Bench-NLP11~\citep{nasbenchx11} were released as surrogate benchmarks that extend NAS-Bench-101, NAS-Bench-301, and NAS-Bench-NLP by predicting the full learning curve information.

\vspace{-2mm}
\section{NAS benchmark statistics} \label{sec:statistics}
\vspace{-1mm}

In this section and in Appendix \ref{app:experiments}, we use \nasbs{} to compute a set of aggregate statistics across a large set of NAS benchmarks. There is a high variance with respect to the distribution of accuracies and other statistics across benchmarks due to substantial differences in the tasks performed and layout of the search space. It is essential to keep this in mind to ensure a fair comparison of the performance of NAS algorithms across these benchmarks. To the best of our knowledge, this is the first large-scale aggregation of statistics computed on NAS benchmarks.

\begin{figure}[t]
    \centering
    \includegraphics[width=\textwidth]{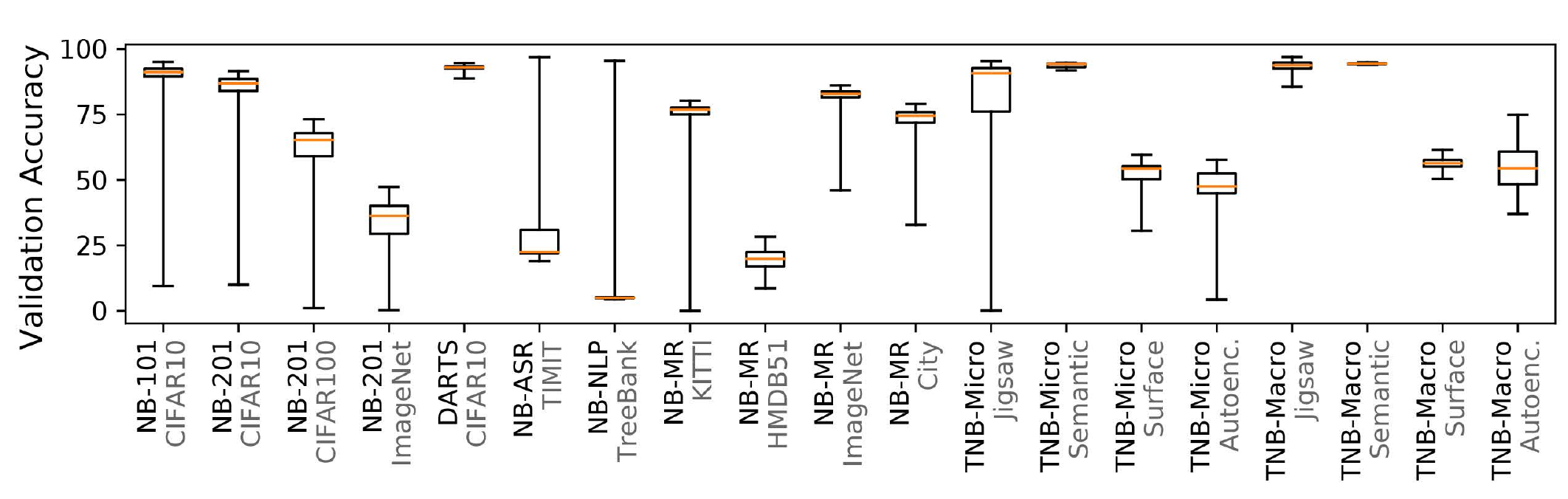}
    \caption{Validation accuracy box plots for each NAS benchmark.
    The whiskers represent the minimum and maximum accuracies in each search space. 
    For NAS-Bench-NLP and TransNAS-Bench, perplexity and SSIM are used instead of validation accuracy, respectively. 
    In the case of extremely large search spaces such as DARTS and NAS-Bench-NLP, the statistics are computed only with respect to the tens-of-thousands of precomputed architectures.
    }
    \label{fig:box_plots}
\end{figure}

\begin{wrapfigure}[21]{r}{.4\textwidth}
\vspace{-3mm}
    \centering
    \includegraphics[width=.4\textwidth]{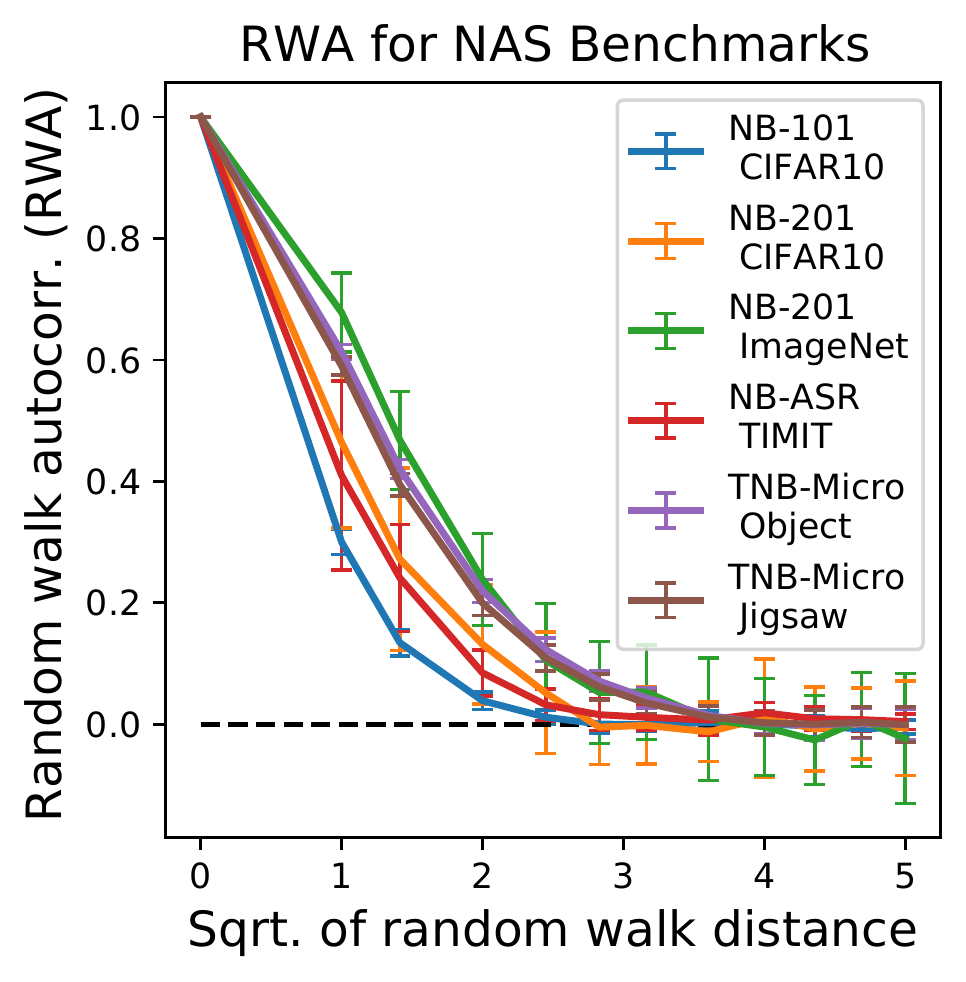}
    \caption{RWA for NAS benchmarks.  
    RWA computes the autocorrelation of accuracies of architectures during a random walk, 
    in which each step perturbs one operation or edge.
    }
    \label{fig:rwa}
\end{wrapfigure}

Figure~\ref{fig:box_plots} shows box plots for the validation accuracy distribution for a representative set of the 25 NAS benchmarks. 
We find that TransNAS-Bench (Sem.\ Segment) and DARTS achieve the highest median and maximum accuracies, yet they also have among the smallest variance in validation accuracy across the search space. 
On the other hand, the search space with the highest interquartile range is TransNAS-Bench Jigsaw.

In Figure~\ref{fig:rwa}, we assess the level of \emph{locality} in each search space, or the similarity of validation accuracy among neighboring architectures (architectures which differ by a single operation or edge) using the random walk autocorrelation (RWA)~\citep{weinberger1990correlated, nasbench, white2021local}. RWA computes the autocorrelation of accuracies of architectures during a random walk, in which each step perturbs one operation or edge. We see that NAS-Bench-201 ImageNet16-120 has the highest autocorrelation, while NAS-Bench-101 has the lowest.
In Appendix~\ref{app:experiments}, we also discuss plots describing the average runtime for training architectures
and the average neighborhood size, for each NAS benchmark. Overall, we see substantial differences among the search spaces along the various axes that we tested.

Overall, we find that the diversity  is important to keep into context when comparing across
many different NAS benchmarks. For example, it is more impressive if a NAS algorithm discovers an architecture within $0.1\%$ of the optimal on NAS-Bench-201 ImageNet16-120, compared to DARTS, because the standard deviation of accuracies for DARTS is much lower.
Additional factors, such as locality and neighborhood size, also affect the difficulty of NAS benchmarks for some NAS algorithms more than for others; for example, locality has a large effect on the performance of regularized evolution but not for random search.

%% file: sections/4_experiments.tex
\vspace{-2mm}
\section{On the Generalizability of NAS Algorithms} \label{sec:experiments}
\vspace{-1mm}

In this section, we carry out a large-scale empirical study on the generalizability of NAS algorithms across diverse search spaces and tasks, using five different black-box algorithms, five different performance predictors, and three one-shot methods across the largest set of NAS benchmarks to date. 
Throughout, we empirically assess three assumptions we have witnessed in the NAS community about the generalizability of NAS algorithms across diverse search spaces and tasks:
\begin{enumerate}[topsep=0pt, itemsep=2pt, parsep=0pt, leftmargin=5mm]
    \item ``If a NAS algorithm does well on the popular NAS benchmarks NAS-Bench-101 and all three datasets of NAS-Bench-201, it surely must generalize to other NAS benchmarks.''
    \item ``NAS algorithms tend to have robust default hyperparameters and do not require tuning.''
    \item ``To improve a NAS algorithm on a new benchmark, we can cheaply optimize its hyperparameters on a tabular benchmark and then transfer the optimized hyperparameters.''
\end{enumerate}

\paragraph{Experimental details.}
A black-box NAS algorithm is an algorithm which iteratively chooses architectures to train, and then uses the final validation accuracies in the next iteration.
We run experiments for five popular black-box NAS algorithms:
random search (RS)~\citep{randomnas, sciuto2019evaluating}, regularized evolution (RE)~\citep{real2019regularized}, local search (LS)~\citep{white2021local, ottelander2020local}, BANANAS~\citep{bananas}, and NPENAS~\citep{npenas}. We run each black-box algorithm for 200 iterations.

Recently, model-based performance prediction methods have gained popularity as subroutines to speed up NAS algorithms \citep{ning2020surgery}. These methods work by training a model using a set of already evaluated architectures, and then using the model to predict the performance of untrained architectures. We compare five popular performance predictors: BOHAMIANN \citep{springenberg2016bayesian}, 
Gaussian process (GP)~\citep{rasmussen2003gaussian}, random forest (RF) \citep{breiman2001random},
neural architecture optimization (NAO) \citep{luo2018neural}, and XGBoost \citep{chen2016xgboost}. We evaluate the performance prediction methods by computing the Spearman rank correlation of the predictions versus ground truth validation accuracy on a held-out test set of 200 architectures.

For each black-box method and each performance predictor, we evaluate the default hyperparameter configuration, as well as 300 randomly sampled hyperparameter configurations, with each reported performance averaged over 10 seeds. 
We give descriptions, implementation details, and hyperparameter details of each method in Appendix \ref{app:experiments}.

\subsection{The Best NAS Methods} \label{subsec:algo}

\begin{figure}[h]
    \centering
    \includegraphics[width=.99\textwidth]{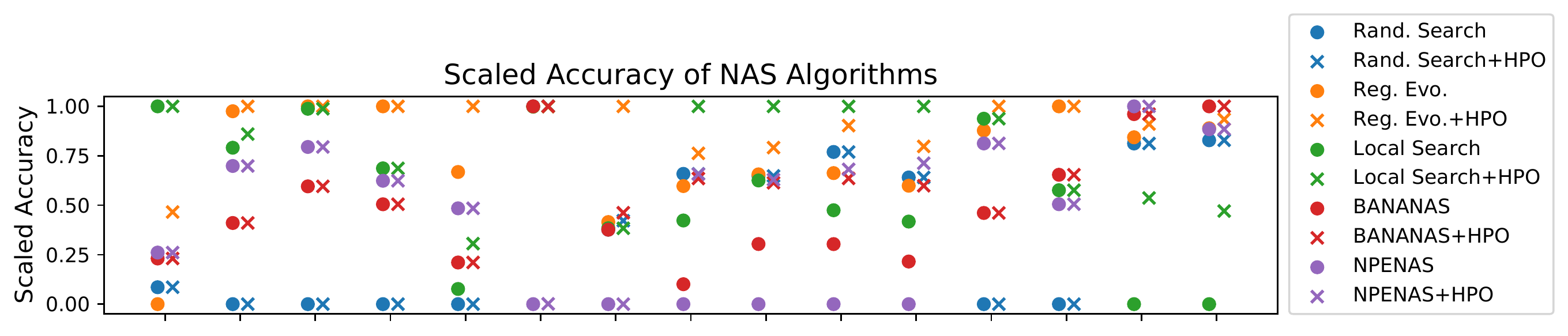}
    \includegraphics[width=.99\textwidth]{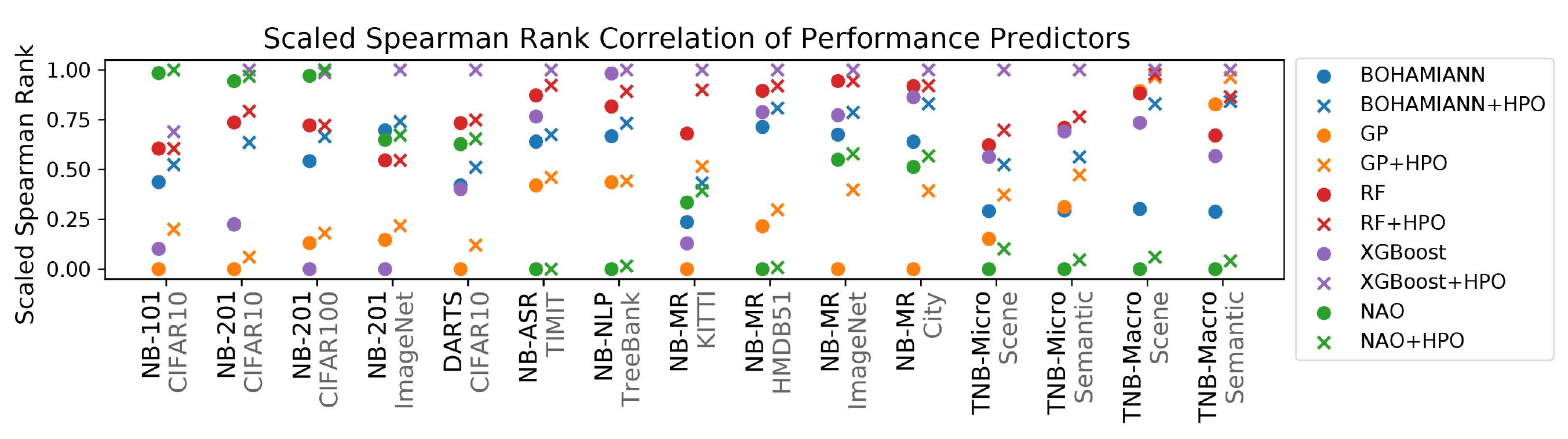}
    \caption{Relative performance of black-box algorithms (top) and performance predictors (bottom) across NAS benchmarks. The solid circles show the performance of the algorithm with default hyperparameters, while the crosses show performance after hyperparameter optimization (HPO).}
    \label{fig:predictors}
\end{figure}

In Figure \ref{fig:predictors}, we plot the scaled (relative) performance of all five black-box algorithms and performance predictors across a representative set of NAS benchmarks (with the full plot in Appendix \ref{app:experiments}). In Table \ref{tab:predictors_summary}, we give a summary by computing the average rank of each black-box algorithm or performance prediction method across all 25 NAS benchmarks.

Across black-box algorithms with their default hyperparameters, we find that no algorithm performs well across all search spaces: no algorithm achieves an average rank close to 1 across all search spaces. RE and LS perform the best on average across the search spaces, with average rankings of 2.36 and 2.66, respectively. 
We also find that although RE performed the best on average, it performs worse than random search in three cases. Therefore, there is no ``best'' black-box algorithm.
When comparing black-box algorithms tuned on each individual benchmark, RE achieves a ranking of 1.96, although we note that since black-box algorithms are expensive to evaluate, it is computationally prohibitive to tune for each individual NAS benchmark.

Across performance predictors, we find that the best predictor with default parameters is RF, and the best predictor when tuned on each individual benchmark is XGBoost, with average rankings of 1.57 and 1.23, respectively. Note that since performance prediction subroutines are often not the bottleneck of NAS, it is common to run hyperparameter tuning during the NAS search. Therefore, we conclude that XGBoost (when tuned) \emph{does} generalize well across all 25 search spaces we tested.

\paragraph{Generalizing beyond NAS-Bench-101 and -201.}
Now we test how well NAS methods generalize from NAS-Bench-101 and -201 to the rest of the NAS benchmarks. In Table \ref{tab:predictors_summary}, for both black-box and predictor methods, we compare the average rank of each method across two different subsets of benchmarks: NAS-Bench-101 and the three different datasets of NAS-Bench-201, versus the rest of the 21 settings excluding NAS-Bench-101 and NAS-Bench-201. We find that for both black-box method and performance predictor methods, the best method \emph{substantially} changes between these two subsets. For example, NAO is the top-performing predictor across NAS-Bench-101 and NAS-Bench-201, yet it achieves very poor performance on the rest of the benchmarks.
This suggests that the insights derived from empirical results are highly dependent on the benchmarks used, and that in order to make reliable claims, evaluating on more than a few benchmarks is crucial.

\begin{table}[h]
\caption{Average relative performance ranking among five NAS algorithms (left) or five performance predictors (right) across 25 settings. Results are weighted by search space; e.g., each of the three NAS-Bench-201 benchmarks are weighted by 1/3.
For abbreviations, see Table \ref{tab:abbreviations}.
}
    \label{tab:predictors_summary}
    
\centering
    \resizebox{\textwidth}{!}{
\begin{tabular}{@{}lcccccccccc@{}}
\toprule
\textbf{} & \multicolumn{5}{c}{\textbf{NAS Algorithms}}                 & \multicolumn{5}{c}{\textbf{Performance Predictors}} \\
          & RS   & RE            & BANANAS & LS   & NPENAS                    & GP    & BOHAM.  & RF             & XGB   & NAO   \\ \midrule
Avg. Rank & 3.47 & \textbf{2.36} & 3.02    & 2.66 & \multicolumn{1}{c|}{3.48} & 4.25  & 3.00       & \textbf{1.57}  & 2.95  & 3.25  \\
Avg. Rank, HPO          & 3.97 & \textbf{1.96} & 3.17 & 2.41          & \multicolumn{1}{c|}{3.49} & 4.37 & 3.36 & 2.41          & \textbf{1.23} & 3.62          \\ \midrule
Avg.Rank, 101\&201      & 4.50 & 3.00          & 3.50 & \textbf{1.50} & \multicolumn{1}{c|}{2.50} & 4.67 & 2.83 & 2.17          & 4.17          & \textbf{1.17} \\
Avg. Rank, non-101\&201 & 3.06 & \textbf{2.11} & 2.83 & 3.13          & \multicolumn{1}{c|}{3.87} & 4.08 & 3.06 & \textbf{1.33} & 2.46          & 4.08          \\ \bottomrule
\end{tabular}
}
\end{table}

\subsection{Generalizability of Hyperparameters} \label{subsec:hpo}

While the previous section assessed the generalizability of NAS methods, now we assess the generalizability of the \emph{hyperparameters} within NAS methods.
For a given NAS method, we can tune it on NAS benchmark $A$, and then evaluate the performance of the tuned method on NAS benchmark $B$, compared to the performance of the best hyperparameters from NAS benchmark $B$. In other words, we compute the ``regret'' of tuning a method on one NAS benchmark and deploying it on another.

\begin{figure}[t]
    \centering
    \includegraphics[width=.49\textwidth]{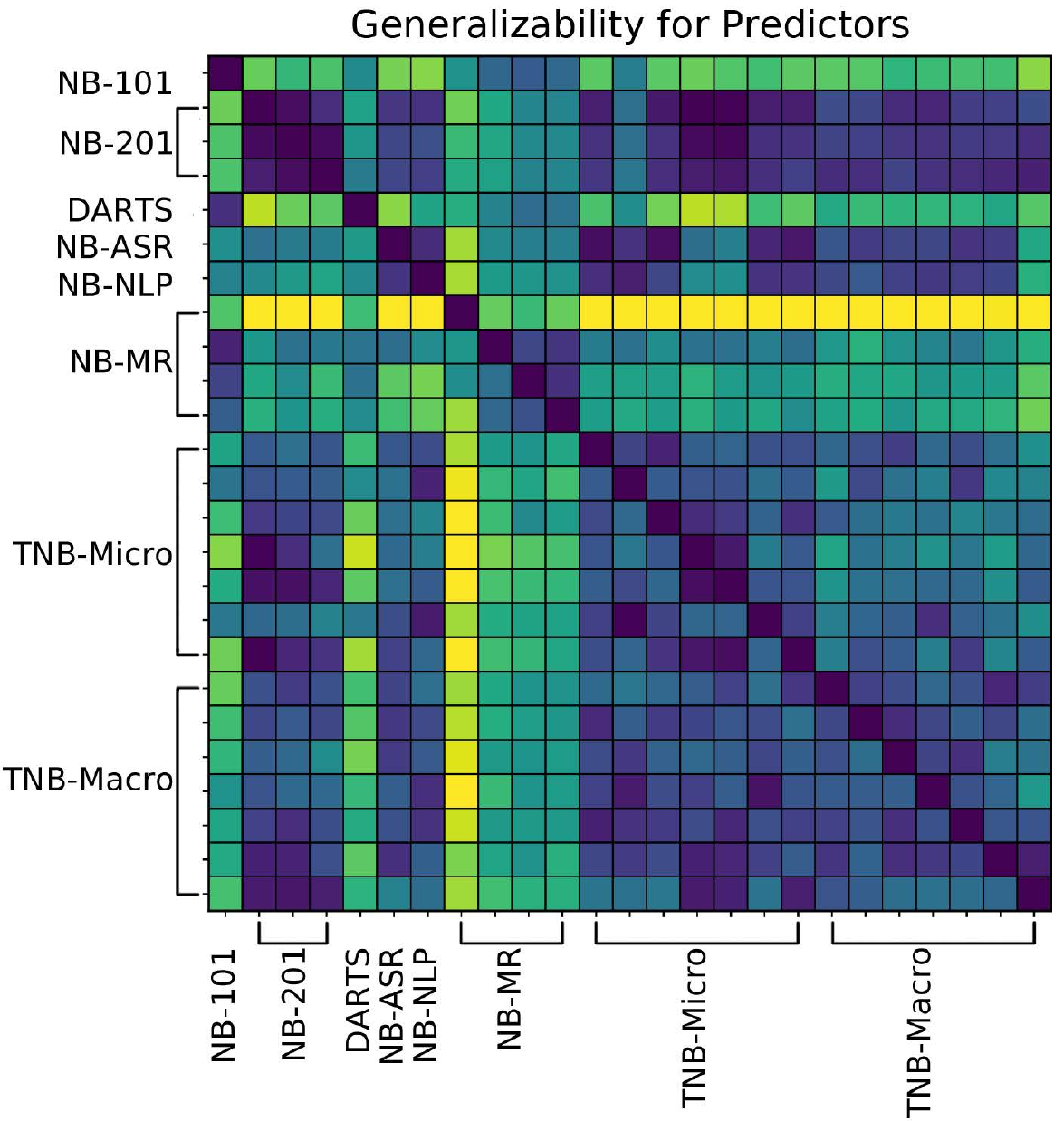}
    \includegraphics[width=.487\textwidth]{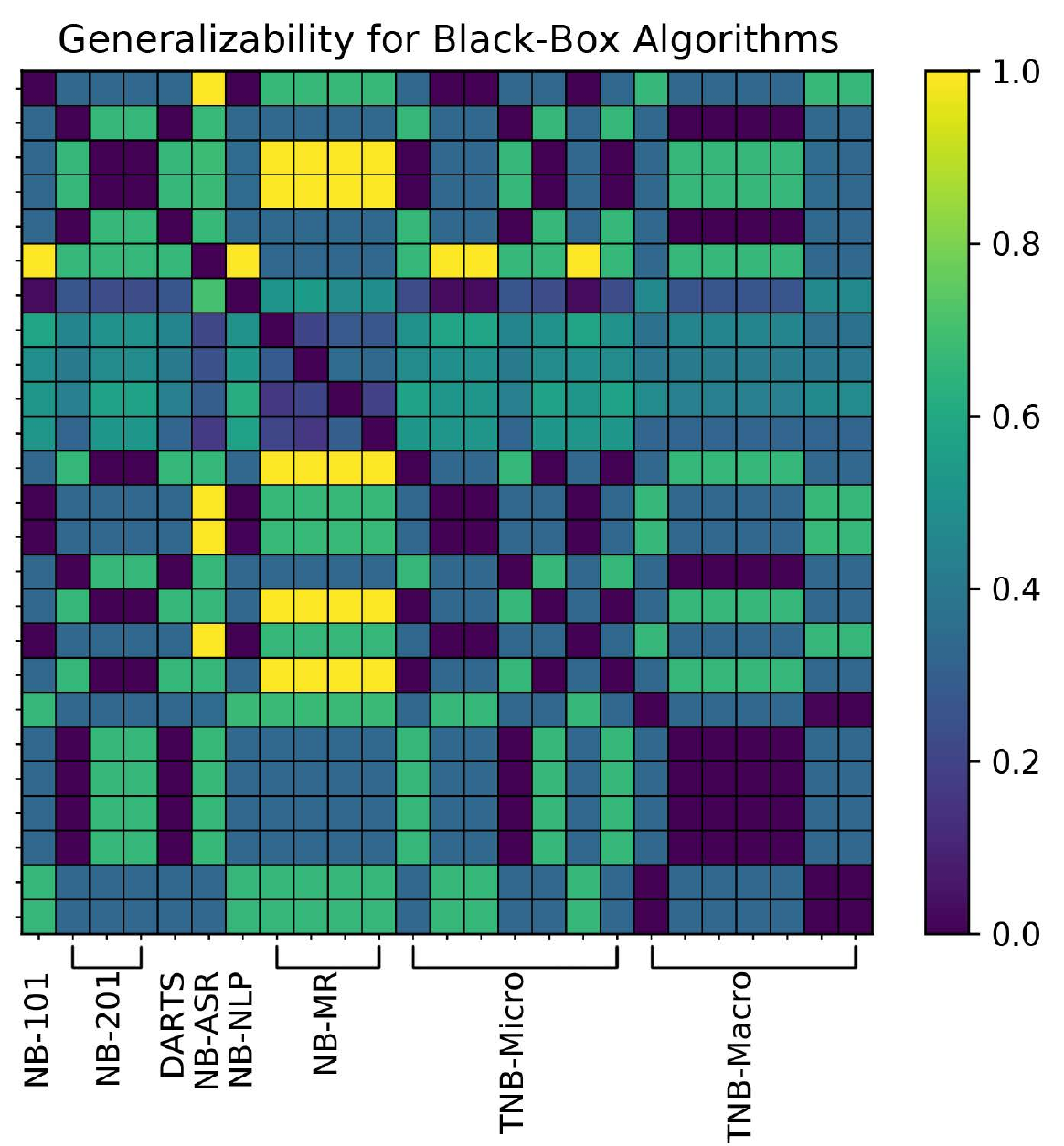}
    \caption{Transferability results for predictors (left) and black-box algorithms (right). 
    Row $i$, column $j$ denotes the scaled regret of an 
    algorithm tuned on search space $i$ and evaluated on search space $j$. 
    For abbreviations, see Table \ref{tab:abbreviations}, and for summary statistics,
    see Appendix \ref{app:algo_experiments}.
    }
    \label{fig:matrices}
    \vspace*{-0.1cm}
\end{figure}

In Figure \ref{fig:matrices} (left), we run this experiment for all pairs of search spaces, averaged over all performance predictors, to give a general estimate of the regret across all search spaces. 
Unsurprisingly, hyperparameters transfer well within a given search space (such as within the three datasets in NAS-Bench-201 or the seven datasets in TransNAS-Bench-Micro). However, we find that no search space achieves strong regret across most search spaces. NAS-Bench-101, DARTS, and the four benchmarks in NAS-Bench-MR have particularly bad regret compared to the other benchmarks.

Next, in Figure \ref{fig:matrices} (right), we run the same experiment for black-box algorithms. We find that the transferability of hyperparameters across black-box algorithms is even worse than across predictors: the hyperparameters do not always transfer well even within different tasks of a fixed search space. We also see that, interestingly, the matrix is less symmetric than for performance predictors. For example, it is particularly hard for hyperparameters to transfer \emph{to} NAS-Bench-MR, but easier to transfer \emph{from} NAS-Bench-MR.
Overall, our experiments show that it is not sufficient to tune hyperparameters on one NAS benchmark and deploy on other benchmarks, as this can often make the performance worse.
In Appendix \ref{app:additional_experiments}, we give additional experiments and summary statistics on the transferability of hyperparameters across search spaces, as well as a guide to interpreting the experiments.
We also present additional experiments that combine our algorithm and statistics experiments to give relationships between properties of the search space and the performance of different algorithms.

\subsection{One-shot Algorithms} \label{subsec:oneshot}
One-shot NAS algorithms, in which a single supernetwork representing the entire search space is trained, are a popular
choice for NAS due to their strong performance and fast runtimes. In this section,
we compare the performance of three one-shot algorithms: DARTS~\citep{darts}, GDAS~\citep{gdas}, and DrNAS~\citep{drnas}, across several different NAS benchmarks. Note that since one-shot algorithms must be able to represent the entire search space in the form of a supernetwork, the algorithms can effectively only be run on cell-based search spaces with a complete graph \citep{nasbench1shot1}, precluding the use of all 25 NAS benchmarks as in the previous section.

\begin{figure}[htp]
    \centering
    \includegraphics[width=.8\textwidth]{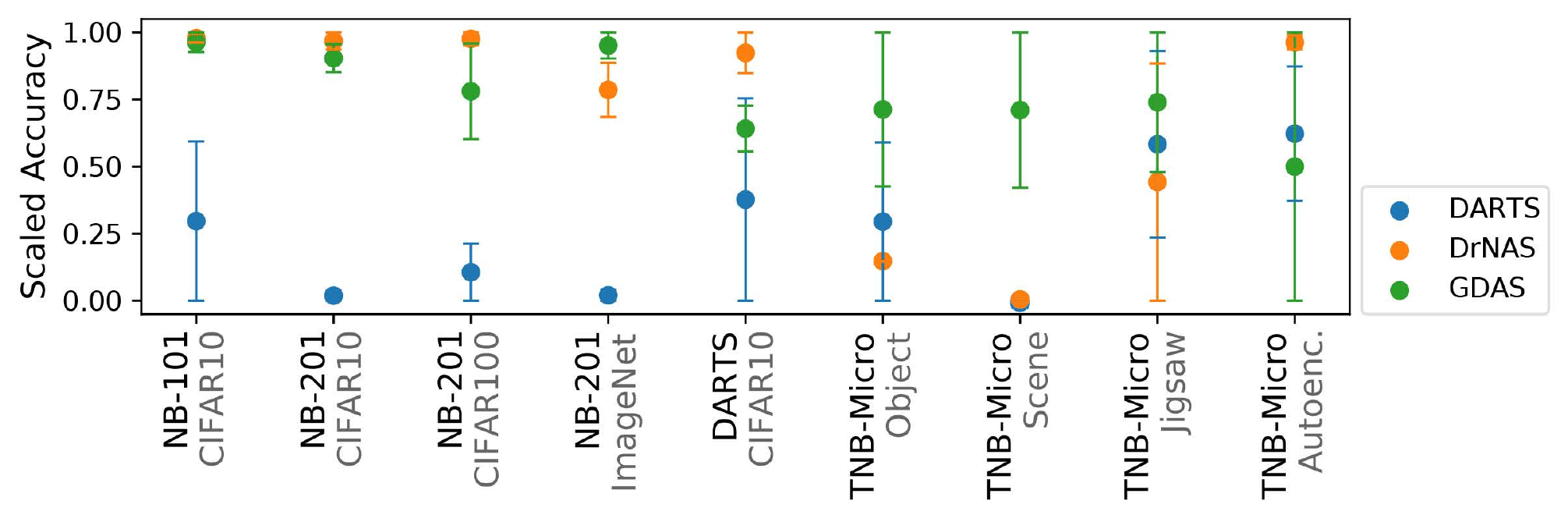}
    \caption{Performance of one-shot algorithms across NAS benchmarks. The bars show the minimum, maximum and average performance over five seeds.
    For abbreviations, see Table \ref{tab:abbreviations}.
    }
    \label{fig:oneshot}
\vspace*{-0.2cm}    
\end{figure}

In Figure~\ref{fig:oneshot}, we plot the scaled (relative) performance of the three one-shot methods run for five seeds each, across nine NAS benchmarks. There is no clear best algorithm: DrNAS performed best on five benchmarks, and GDAS performed best on the remaining four. DARTS did not perform as well, which is consistent with prior work \citep{zela2020understanding, nasbench201}.

Throughout this section, we showed that many implicit assumptions in the NAS community regarding NAS algorithm generalizability are incorrect, and that it is important to consider a large set of NAS benchmarks in order to avoid false conclusions. In the next section, in order to help NAS researchers and practitioners avoid these pitfalls, we describe our new NAS benchmark suite, which is designed with the aim to help the community develop generalizable NAS methods.

%% file: sections/5_codebase.tex
\vspace{-2mm}
\section{\nasbs: Overview} \label{sec:codebase}
\vspace{-1mm}

In this section, we give an overview of the \nasbs{} codebase, which allows researchers to easily compare NAS algorithms across many tasks, as shown in Section \ref{sec:experiments}.

\subsection{Key challenges in creating a flexible benchmark suite}

Before the introduction of tabular NAS benchmarks, it was common practice to release new NAS methods along with custom-designed search spaces to go with the methods~\citep{lindauer2019best, randomnas}, leading to many early NAS methods being intertwined with and hard to separate from their original search space. This was especially true for weight sharing methods, many of which require specific constraints on the topology of the NAS structure and specific sets of operations in order to run properly. The result was that many algorithms needed substantial code changes in order to run on other search spaces.

\begin{wrapfigure}[18]{r}{8.3cm}
\vspace{-5mm}
\begin{lstlisting}[caption={A minimal example on how one can run a NAS algorithm in \nasbs{}. Both the search space and the algorithm can be changed in one line of code.},label=fig:one-line,language=Python]
from naslib.search_spaces import NasBench101SearchSpace
from naslib.optimizers import RegularizedEvolution
from naslib.defaults.trainer import Trainer
from naslib.utils import utils, get_dataset_api

config = utils.get_config_from_args(config_type='nas')

search_space = NasBench101SearchSpace()
optimizer = RegularizedEvolution(config)

dataset_api = get_dataset_api(config.search_space, 
                              config.dataset)

optimizer.adapt_search_space(search_space, 
                             dataset_api=dataset_api)
trainer = Trainer(optimizer, config)
trainer.search()
trainer.evaluate()
\end{lstlisting}
\end{wrapfigure}
Even after the release of several queryable benchmarks, it is still not common practice to run NAS algorithms on more than a few benchmarks due to the nontrivial differences among each benchmark.
For example, as described in Section \ref{sec:nasbench_overview}, operations on the nodes~\citep{nasbench} versus on the edges~\citep{darts} 
added complexity in adapting one-shot optimizers to many search spaces, and for some search spaces one-shot optimizers could only be run on subsets of the full space~\citep{nasbench1shot1}. Other differences, such as the presence of hidden nodes~\citep{nasbenchnlp} or skip connections~\citep{nasbenchasr},
cause NAS components to require different implementations.
Creating a robust NAS benchmark suite is not as simple as ``combining the individual codebases'', because such a solution would require re-implementing each new NAS algorithm on several search spaces.
In fact, the difficulty in creating a benchmark suite is likely a key reason why there are so few papers that evaluate on more than a few benchmarks.

\subsection{The \nasbs{} codebase}
To overcome these difficulties, \nasbs{} enacts two key principles: flexibility in defining the search space and modularity of individual NAS components. We first describe how we achieve flexible search space definitions, and then we detail the modular design of \nasbs{}.

A search space is defined with a graph object using PyTorch and NetworkX~\citep{networkx}, a package that allows for easy-to-use and flexible graph creation and manipulations.
This functionality allows for a dynamic search space definition, where candidate operations can be encoded both as part of a node ~\citep{nasbench, nasbenchnlp} as well as a edge ~\citep{nasbench201, darts}.
It also allows the representation of multiple layers of graphs on top of the computational graph, allowing the formation of nested graph-structures that can be used to define hierarchical spaces ~\citep{ru2020neural, liu2019auto}.

The \nasbs{} is modular in the sense that individual NAS components, such as the search space, NAS algorithm, or performance predictors, are disentangled and defined separately from one another. In Snippet~\ref{fig:one-line}, we showcase a minimal example that runs a black-box NAS algorithm on a tabular benchmark in \nasbs{}. Other optimizers and benchmarks can be imported and run similarly.
Due to these design principles, \nasbs{} allows researchers to implement their NAS algorithm in isolation, and then evaluate on all the benchmarks integrated in \nasbs{} without writing any additional code.
Since a variety of NAS algorithms, search spaces, and performance predictors have already been integrated into the open-source framework, this allows for the user to build on top of predefined NAS components.
With the entire pipeline in place, along with the possibility of quick evaluations across search spaces and tasks, we believe that \nasbs{} will allow researchers to rapidly prototype and fairly evaluate NAS methods. All the scripts to run the evaluations conducted in this paper come together with the library codebase. For more details on the API, see Appendix \ref{app:codebase}.

%% file: sections/8_related_work.tex
\vspace{-2mm}
\section{Related Work}
\vspace{-1mm}

We describe work that provides experimental surveys, benchmark suites, or unified codebases within NAS. For detailed surveys on NAS, see~\citep{nas-survey,xie2020weight}.

\noindent\textbf{NAS experimental surveys.}
Multiple papers have found that random search is a competitive NAS 
baseline~\citep{randomnas, sciuto2019evaluating}, including a recent work that benchmarked eight NAS methods with five datasets on the DARTS search space \citep{yang2019evaluation}.
Other recent works have shown experimental surveys for NAS performance 
predictors~\citep{ning2020surgery, white2021powerful},
and experimental analyses on weight sharing~\citep{yu2020train}.

\noindent\textbf{Benchmark suites.}
NAS-Bench-360 \citep{nasbench360} is a very recent benchmark suite which presents NAS benchmarks for ten diverse datasets on three search spaces. 
However, a drawback is that evaluating NAS algorithms requires 1 to $100+$ GPU-hours \citep{nasbench360}. This is in contrast to the \nasbs{}, where NAS algorithms take at most 5 minutes on a CPU due to the use of queryable benchmarks. Outside of NAS, multiple hyperparameter tuning benchmark suites have been released \citep{hpobench, hpob}.

\noindent\textbf{NAS codebases.}
The \emph{DeepArchitect} library~\citep{negrinho2017deeparchitect} was the first to have a modular design for search spaces and NAS algorithms.
\emph{PyGlove}~\citep{peng2021pyglove} is a library for NAS featuring dynamically adapting components, however, it is not open-sourced.
\emph{Neural Network Intelligence} (NNI) \citep{nni}
is a platform for AutoML that implements many algorithms as well as NAS-Bench-101 and NAS-Bench-201.
Other NAS repositories are actively being built, such as
\emph{archai} \citep{archai} and \emph{aw\_nas} \citep{ning2020surgery}.

%% file: sections/9_conclusion.tex
\vspace{-2mm}
\section{Conclusion and Future Work} \label{sec:conclusion}
\vspace{-1mm}

In a large-scale study across 25 NAS benchmark settings, 5 blackbox NAS methods, 5 NAS predictors, and 3 one-shot methods, we showed that many implicit assumptions in the NAS community are wrong. 
Firstly, there is no single best NAS method: which method performs best very much depends on the benchmark. Along similar lines, if a NAS method performs well on the popular NAS benchmarks NAS-Bench-101 and all three datasets of NAS-Bench-201, in contrast to what one might have expected, this does \emph{not} imply that it will also perform well on other NAS benchmarks.
Finally, tuning a NAS algorithm's hyperparameters can make it dramatically better, but transferring such hyperparameters across benchmarks often fails.
This analysis strongly suggests adapting the empirical standard of the field, to stop focusing too much on smaller NAS benchmarks like NAS-Bench-101 and NAS-Bench-201, and rather also embrace larger and novel NAS benchmarks for natural language processing~\citep{nasbenchnlp}, automatic speech recognition~\citep{nasbenchasr}, and pixel-level prediction, regression, and self-supervised tasks~\citep{transnasbench}.

While substantial differences across NAS search spaces has so far made it very hard to use many NAS benchmarks, we showed a way out of this dilemma by introducing an easy-to-use, unified benchmark suite that we hope will facilitate reproducible, generalizable, and rapid NAS research.

For future work, \nasbs{} can benefit from additional options, such as distributed training.
Furthermore, although practitioners using NAS-Bench-Suite have the option to choose their own hand-picked subset of the 25 tasks based on their specific application, it would be useful to define representative subsets of the benchmarks in NAS-Bench-Suite based on application type.

%% file: sections/10_ethics_statement.tex
\section{Ethics Statement} \label{sec:impact}


Our work gives a large scale evaluation of generalizability in NAS and then
proposes a new benchmark suite for NAS. The goal of our work is to make it
faster and more accessible for researchers to run generalizable and reproducible NAS experiments.
Specifically, the use of tabular and surrogate NAS benchmarks allow researchers to
simulate NAS experiments cheaply on a CPU, rather than requiring a GPU cluster, reducing the
carbon footprint of NAS research~\citep{patterson2021carbon, hao2019training}.
This is especially important since the development stage of NAS research may be extremely
computationally intensive without the use of NAS benchmarks~\citep{zoph2017neural,real2019regularized}.
Our work is a tool for the NAS community, which facilitates NAS research that may be used for
positive impacts on society (for example, algorithms that reduce $\text{CO}_2$ 
emissions~\citep{rolnick2019tackling}) or negative impacts on society (for example, models that
discriminate or exclude groups of people).
Due to the increase in conversations about ethics and societal impacts in the AI 
community~\citep{hecht2018time}, we are hopeful that the applications of our work
will have a net positive impact on society.

%% file: sections/11a_nas_reproducibility.tex
\section{Best Practices for NAS Research}\label{app:nas_checklist}
There have been a few recent works which have called for improving the reproducibility and 
fairness in experimental comparisons in NAS research~\citep{randomnas, nasbench, yang2019evaluation}.
This led to the release of a NAS best practices checklist~\citep{lindauer2019best}.
We address each part of the checklist.

\paragraph{Best practices for releasing code.}
For each NAS benchmark that we used, the code for the training pipeline and the search space
is already publicly available.
Since we used NAS benchmarks for all of our experiments, we did not evaluate the
architectures ourselves. 
All of the code for the NAS methods, including the hyperparameters, are available at \url{https://github.com/automl/naslib}. We discuss the choices of hyperparameters in Appendix \ref{app:experiments}.   

\paragraph{Best practices for comparing NAS methods.}
Since we made use of NAS benchmarks, all of the details for training the architectures are fixed across NAS methods. We included baselines such as random search and local search in our experiments in Section \ref{sec:experiments}. We averaged 20 trials and 100 or more hyperparameter configurations for each experiment, and the choice of seeds (0-19) and hyperparameter configurations is available at \url{https://github.com/automl/naslib}.

\paragraph{Best practices for reporting important details.}
We reported how we tuned the hyperparameters of NAS methods in Section \ref{sec:experiments}.
We included all details of our experimental setup in Section \ref{sec:experiments} and \ref{app:experiments}.

%% file: sections/A_appendix_benchmarks.tex
\section{Details from Section~\ref{sec:nasbench_overview}}\label{sec:appendix_nasbench_overview}

\begin{table}[t]
\centering
\caption{List of abbreviations used in the text.}
\begin{tabular}{@{}ll@{}}
\toprule
Abbreviation & Full                                                             \\ \midrule
RS           & Random Search                                                    \\
RE           & Regularized Evolution                                            \\
LS           & Local Search                                                     \\
NPENAS       & Neural Predictor Guided Evolution for Neural Architecture Search \\
BANANAS   & Bayesian Optimization with Neural Architectures for Neural Architecture Search                                          \\ \midrule
GP           & Gaussian Process                                                 \\
NAO          & Neural Architecture Optimization                                 \\
RF           & Random Forest                                                    \\
XGB          & XGBoost (Extreme Gradient Boosting)                              \\
BOHAMIANN & \begin{tabular}[c]{@{}l@{}}Bayesian Optimization with Hamiltonian Monte Carlo Artificial Neural\\ Networks\end{tabular} \\ \midrule
NB-101       & NAS-Bench-101                                                    \\
NB-201       & NAS-Bench-201                                                    \\
NB-301       & NAS-Bench-301                                                    \\
NB-ASR       & NAS-Bench-ASR (Automated Speech Recognition)                     \\
NB-NLP       & NAS-Bench-NLP (Natural Language Processing)                      \\
NB-MR        & NAS-Bench-MR (Multi-Resolution)                                  \\
TNB-Micro    & TransNAS-Bench, Micro search space                               \\
TNB-Macro    & TransNAS-Bench, Macro search space                               \\ \midrule
DARTS        & Differentiable Architecture Search                               \\
GDAS         & Gradient-based search using Differentiable Architecture Sampler  \\
DrNAS        & Dirichlet Neural Architecture Search                             \\ \bottomrule
\end{tabular}
\label{tab:abbreviations}
\end{table}

This section is an extension of the ``NAS benchmarks'' part of Section~\ref{sec:nasbench_overview}, with additional details for each NAS benchmark, as well as a more comprehensive list of NAS benchmarks. For an extension of Table \ref{tab:benchmarks}, which contains a more comprehensive list, see Table \ref{tab:benchmarks_comprehensive}.

\paragraph{NAS benchmarks.}
The first tabular NAS benchmark to be released was NAS-Bench-101~\citep{nasbench}.
This benchmark consists of a cell-based search space of $423\,624$ architectures
with a fixed macro structure. NAS-Bench-101 comes with precomputed 
validation and test accuracies at epochs 4, 12, 36, and 108 from training on CIFAR-10.
The cell-based search space of NAS-Bench-101 consists of five nodes which can take on 
any DAG structure with at most seven edges. Each node can take on one of three operations.
Since training with stochastic gradient descent is random, 
all architectures were trained three times with different seeds and therefore
have three sets of accuracies.

Since NAS-Bench-101 architectures contain a variable amount of nodes, it is not
possible to evaluate one-shot algorithms. Therefore, NAS-Bench-1Shot1~\citep{nasbench1shot1}
defines three subsets of NAS-Bench-101 which allow one-shot algorithms to be run.
The largest subset size in NAS-Bench-1Shot1 is $363\,648$.

NAS-Bench-201 is the second tabular NAS benchmark. It consists of a cell
which is a complete directed acyclic graph over 4 nodes. 
Therefore, there are $\binom{4}{2}=6$ edges.
Each edge can take on one of five operations (note that this is in contrast to
NAS-Bench-101, in which the \emph{nodes} are operations).
The search space consists of $5^6=15,625$ neural architectures,
although due to \texttt{none} and \texttt{identity} operations, the number
of non-isomorphic architectures is $6\,466$.
Each architecture has precomputed train, validation, and test losses and accuracies
for 200 epochs on CIFAR-10, CIFAR-100, and ImageNet-16-120.
As in NASBench-101, on each dataset, each architecture was trained three times
using different random seeds.

NATS-Bench~\citep{natsbench} is an extension of NAS-Bench-201 which also varies
the macro architecture. Specifically, a search space of $32\,768$ architectures
with varying size were trained across three datasets for three seeds.

The first non computer vision NAS benchmark to be released was NAS-Bench-NLP~\citep{nasbenchnlp}.
Its search space consists of a DAG of up to 24 nodes, each of which can take on one
of seven operations and can have at most three incoming edges. With a size of at least
$10^{53}$, NAS-Bench-NLP is currently the largest NAS benchmark.
$14\,322$ of the architectures were trained on Penn Tree Bank~\citep{penntreebank} for 50 epochs.
Since only a fraction of architectures were trained, NAS-Bench-NLP is not queryable.

The DARTS~\citep{darts} search space with CIFAR-10 is arguably the most popular NAS benchmark.
The search space contains $10^{18}$ architectures, consisting 
of two cells, each of which has six nodes. Each node has exactly two incoming edges,
and each edge can take one of eight operations.
Recently, $60\,000$ of the architectures were trained for 100 epochs and used to create
NASBench-301~\cite{nasbench301}, the first surrogate NAS benchmark. The authors released
pretrained surrogates created using XGBoost~\cite{chen2016xgboost} and graph isomorphism networks~\citep{xu2018powerful}.

NAS-Bench-ASR~\citep{nasbenchasr} is a tabular NAS benchmark for automatic
speech recognition. The search space consists of $8\,242$ architectures trained on the TIMIT
dataset. The search space consists of four nodes, with three main edges that can take on
one of six operations, and six skip connection edges, which can be set to on or off.

NAS-Bench-111, NAS-Bench-311, and NAS-Bench-NLP11~\citep{nasbenchx11} were recently released as surrogate benchmarks
that extend NAS-Bench-101, NAS-Bench-301, and NAS-Bench-NLP by predicting the full learning curve information.
In particular, none of NAS-Bench-101, NAS-Bench-301, and NAS-Bench-NLP allow the validation accuracies to be
queried at arbitrary epochs, which is necessary for multi-fidelity NAS techniques such as 
learning curve extrapolation~\citep{baker2017accelerating, lcnet}.
The surrogates used to create NAS-Bench-111, NAS-Bench-311, and NAS-Bench-NLP11 include singular value
decomposition and noise modeling~\citep{nasbenchx11}.

TransNAS-Bench~\citep{transnasbench} is a tabular NAS benchmark consisting of two separate search spaces (cell-level and macro-level) and seven tasks including pixel-level prediction, regression, and self-supervised tasks. The-cell level search space of TransNAS-Bench is similar to that of NAS-Bench-201, but with 4 choices of operations per edge, hence $4\,096$ architectures altogether. The macro-level search space instead has a flexible macro skeleton with variable number of blocks, locations to down-sample feature maps, and locations to raise the channels, leading to a total of $3\,256$ architectures.

NAS-Bench-MR \citep{nasbenchmr} is a surrogate NAS benchmark which evaluates nine settings total, across four datasets: ImageNet50-1000, Cityscapes, KITTI, and HMDB51. NAS-Bench-MR consists of a single search space of size $10^{23}$, and for each of the nine settings, $2\,500$ architectures were trained, to create nine different surrogates for each of the nine settings.

NAS-Bench-360 \citep{nasbench360} is a very recent benchmark suite which gives NAS benchmarks for ten different datasets, including tasks that are novel for NAS such as spherical projection, fluid dynamics, DNA sequencing, medical imaging, surface electromyography, and cosmic ray detection. The tasks are carried out on three different search spaces based on Wide ResNet \citep{resnet}, DARTS \citep{darts}, and DenseNAS \citep{densenas}. However, a drawback of NAS-Bench-360 is that none of the NAS benchmarks are queryable. Therefore, evaluating NAS algorithms on these benchmarks requires 1 to $100+$ GPU-hours of runtime \citep{nasbench360}.

NAS-Bench-Macro \citep{nasbenchmacro} is a NAS benchmark which focuses on the macro search space. It consists of 6561 pretrained architectures on CIFAR-10. The search space consists of 8 layers, each with 3 choices of blocks.

HW-NAS-Bench is a NAS benchmark focusing on hardware-aware neural architecture search. It gives the measured/estimated hardware-cost for all architectures in NAS-Bench-201 and FBNet \citep{fbnet} on six hardware devices, including commercial edge, FPGA, and ASIC devices. HW-NAS-Bench can be used alongside NAS-Bench-201 for the full information on hardware cost and model accuracy for all architectures in NAS-Bench-201.

\begin{table}[t]
\caption{A comprehensive overview of NAS benchmarks.}
\resizebox{\linewidth}{!}{%
\centering
\begin{tabular}{@{}lcccccccc@{}}
\toprule
\multicolumn{1}{l}{} & \multicolumn{1}{c}{} & \multicolumn{2}{c}{\textbf{Queryable}} 
& \multicolumn{1}{c}{} & \multicolumn{1}{c}{} & \multicolumn{1}{c}{} & \multicolumn{1}{c}{} \\
\cmidrule{3-4} \textbf{Benchmark} & \textbf{Size} & \textbf{Tab.} & \textbf{Surr.} 
& \textbf{LCs} & \textbf{Macro} & \textbf{Type} & \textbf{\#Tasks} & \textbf{NAS-Bench-Suite} \\
\midrule 
NAS-Bench-101 & 423k & \cmark & & & & Image\ class.\ & 1 & \cmark \\
\midrule 
NAS-Bench-201 & 6k   & \cmark & & \cmark &  & Image\ class.\ & 3 & \cmark \\
\midrule 
NATS-Bench & 6k   & \cmark & & \cmark & \cmark & Image\ class.\ & 3 & \cmark \\
\midrule
NAS-Bench-NLP & $10^{53}$ & &  & \cmark &  & NLP & 1 & \cmark \\
\midrule 
NAS-Bench-1Shot1 & 364k & \cmark & & & & Image\ class.\ & 1 & \cmark \\
\midrule 
NAS-Bench-301 & $10^{18}$ & & \cmark & & & Image\ class.\ & 1 & \cmark \\
\midrule
NAS-Bench-ASR & 8k & \cmark & & & \cmark & ASR & 1 & \cmark \\
\midrule
TransNAS-Bench & 7k & \cmark & & \cmark & \cmark & Var.\ CV & 14 & \cmark \\
\midrule
NAS-Bench-111 & 423k & & \cmark & \cmark & & Image\ class.\ & 1 & \cmark \\
\midrule
NAS-Bench-311 & $10^{18}$ & & \cmark  & \cmark & & Image\ class.\ & 1 & \cmark \\
\midrule
NAS-Bench-NLP11 & $10^{53}$ & & \cmark & \cmark & & NLP & 1 & \cmark \\
\midrule
NAS-Bench-MR & $10^{23}$ & & \cmark & & \cmark & Var.\ CV & 9 & \cmark \\
\midrule
NAS-Bench-360 & Var.\ & & & & \cmark & Var.\ & 30 & \\
\midrule
NAS-Bench-Macro & 6k & \cmark & & & \cmark & Image\ class.\ & 1 & \\
\midrule
HW-NAS-Bench (201) & 6k & \cmark & & \cmark & & Image\ class.\ & 3 & \\
\midrule
HW-NAS-Bench (FBNet) & $10^{21}$ & & & & & Image\ class.\ & 1 & \\
\bottomrule
\end{tabular}
}
\label{tab:benchmarks_comprehensive}
\end{table}

%% file: sections/B_appendix_experiments.tex
\section{Details from Section~\ref{sec:experiments}} \label{app:experiments}

\subsection{NAS algorithm implementation details}

Here, we give implementation details for all algorithms that we compared in 
Section~\ref{sec:experiments}.
We made an effort to keep the implementations as close as possible
to the original implementation. We start with the black-box optimizers.
For a list of the default hyperparameters and hyperparameter ranges, see \url{https://github.com/automl/NASLib}.

\begin{itemize}
    \item\textbf{Random search.} 
    Random search is the simplest baseline for NAS \cite{randomnas, sciuto2019evaluating}. 
    It draws architectures at random and then returns the best architecture.

    \item\textbf{Local search.} 
    Another baseline, local search has been shown to perform 
    well on multiple NAS benchmarks~\citep{white2021local, ottelander2020local, nasbench301}. 
    It works by evaluating all
    architectures in the neighborhood of the current best architecture found so far.
    The neighborhood of an architecture is the set of architectures which differ by
    one operation or edge.
    We used the implementation from White et al.~\citep{white2021local}.
    
    \item\textbf{Regularized evolution.} 
    This algorithm~\citep{real2019regularized} consists of iteratively mutating the best architectures drawn from a sample of the most recent architectures evaluated. A mutation is defined by randomly changing one operation or
    edge. We used the NAS-Bench-101~\citep{nasbench} implementation.

    \item \textbf{BANANAS.} This NAS algorithm~\citep{bananas} 
    uses Bayesian optimization with an ensemble of three predictors as the surrogate.
    We use the code from the original repository, but using PyTorch for the MLPs instead of Tensorflow. We use the adjacency matrix encoding instead of the path encoding, since the path encoding does not scale to large search spaces such as NAS-Bench-NLP. We use variational sparse GPs \citep{titsias2009variational} in the ensemble of predictors, since this was shown in prior work to perform well and have low runtime~\citep{white2021powerful}.
    
    \item \textbf{NPENAS.} This algorithm \citep{npenas} is based on predictor-guided evolution. It iteratively chooses the next architectures by mutating the most recent architectures in a random sample of the population to create a set of candidate architectures, and then using a predictor to pick the architectures with the highest expected accuracy. Again, we use variational sparse GP \citep{titsias2009variational} as the predictor.

\end{itemize}

Now we describe the performance predictors. For each method, we used the adjacency one-hot encoding \citep{white2020study}.

\begin{itemize}
    \item \textbf{BOHAMIANN.}
    BOHAMIANN~\citep{springenberg2016bayesian} is a Bayesian inference prediction method which uses
    stochastic gradient Hamiltonian Monte Carlo (SGHMC) in order to sample from a Bayesian Neural Network. We use the original implementation from the \texttt{pybnn} package.
    \item \textbf{GP.}
    Gaussian Process (GP)~\citep{rasmussen2003gaussian} is a popular surrogate often used with Bayesian optimization \citep{frazier2018tutorial,snoek2012practical}. Every feature has a joint Gaussian distribution. We use the \texttt{Pyro} implementation~\citep{bingham2019pyro}.
    \item \textbf{NAO.}
    Neural Architecture Optimization is a NAS method that uses an encoder-decoder~\citep{luo2018neural}. The encoder is a feedfoward neural network, and the decoder is an LSTM and attention mechanism. We used the implementation from SemiNAS~\citep{seminas}.
    \item \textbf{Random Forest.} Random forests~\citep{breiman2001random}
    are ensembles of decision trees. Random forests have been used as 
    model-based predictors in NAS~\citep{nasbench301,white2021powerful}.
    We use the Scikit-learn implementation~\citep{pedregosa2011scikit}.
    \item \textbf{XGBoost.} eXtreme Gradient Boosting (XGBoost)~\citep{chen2016xgboost}
    is a popular gradient-boosted decision tree which has been used in NAS \citep{nasbench301,white2021powerful}. We used the original code \citep{chen2016xgboost}.
    
\end{itemize}

For one-shot methods, we mainly focus on the differentiable architecture search approaches, and select popular algorithms like DARTS~\citep{darts}, GDAS~\citep{gdas} and DrNAS~\citep{drnas} in our paper.

\begin{itemize}
    \item \textbf{DARTS.} DARTS~\citep{darts} is the first work on differentiable architecture search. Compared to normal one-shot method that has a one-hot encoding to select one architecture out of different choices, it uses a vector, a.k.a. architecture parameters, where each element ranges from 0 to 1 as probability. During training, it sums all branches as a weight summation. After the search, the final architecture is converted by selecting the branch that with highest weight.
    \item \textbf{GDAS.} Since differentiable architecture search suffers from unstable training compared to traditional one-shot methods, GDAS~\citep{gdas} bridges the gap. Instead of having a weighted summation of all paths, it discretizes the paths during training to select the path with highest probability, while the rest of algorithms remains similar as original DARTS.
    \item \textbf{DrNAS.} Dirichlet architecture search (DrNAS)~\citep{drnas} is another attempt to solve the instability issue of differentiable architecture search. This work treats the continuously relaxed architecture weights as a random variable, which is modeled by a Dirichlet distribution. To this end, the Direchlet parameters can be updated by the traditional differentiable architecture search optimizer easily in the end-to-end manner. 
\end{itemize}


\section{Additional Experiments} \label{app:additional_experiments}

In this section, we give additional statistics, algorithm, and insight experiments to augment Sections \ref{sec:statistics} and \ref{sec:experiments}.

\subsection{Additional Statistics Experiments} \label{app:stats_experiments}

\begin{figure}[t]
    \centering
    \includegraphics[width=\textwidth]{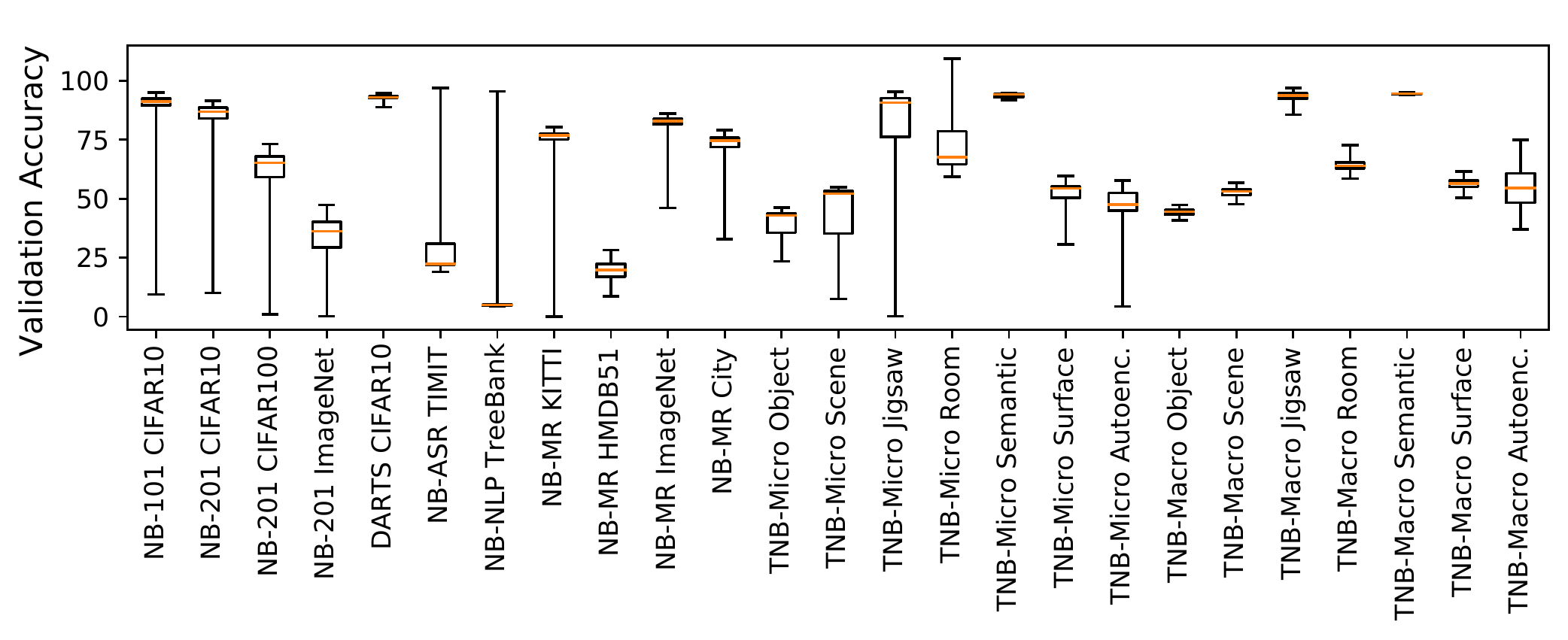}
    \caption{
    Validation accuracy box plots for each NAS benchmark.
    The whiskers represent the minimum and maximum accuracies in each search space. 
    For NAS-Bench-NLP, perplexity is used instead of validation accuracy, and three datasets of TransNAS-Bench do not use accuracy: Surface Normal uses SSIM, Autoencoding uses SSIM, and Room Layout uses negative loss. These are in accordance with the metrics used in the original work. 
    Finally, in the case of extremely large search spaces such as DARTS and NAS-Bench-NLP, the statistics are computed only with respect to the tens-of-thousands of precomputed architectures.
    }
    \label{fig:box_plots_appendix}
\end{figure}

\begin{figure}[t]
    \centering
    \includegraphics[width=\textwidth]{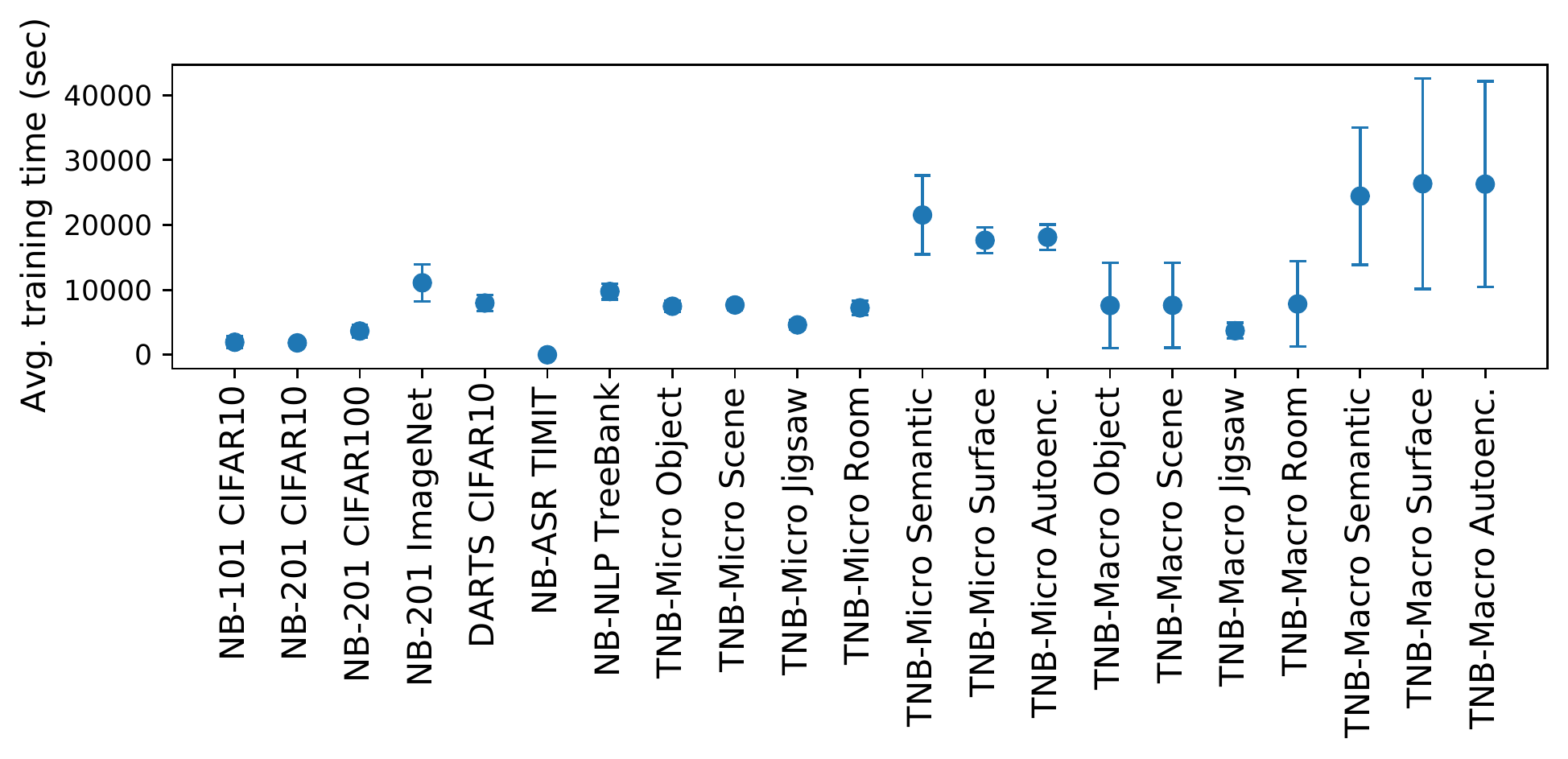}
    \caption{Average time to train an architecture for each NAS benchmark.
    }
    \label{fig:runtimes}
\end{figure}

\begin{figure}[t]
    \centering
    \includegraphics[width=\textwidth]{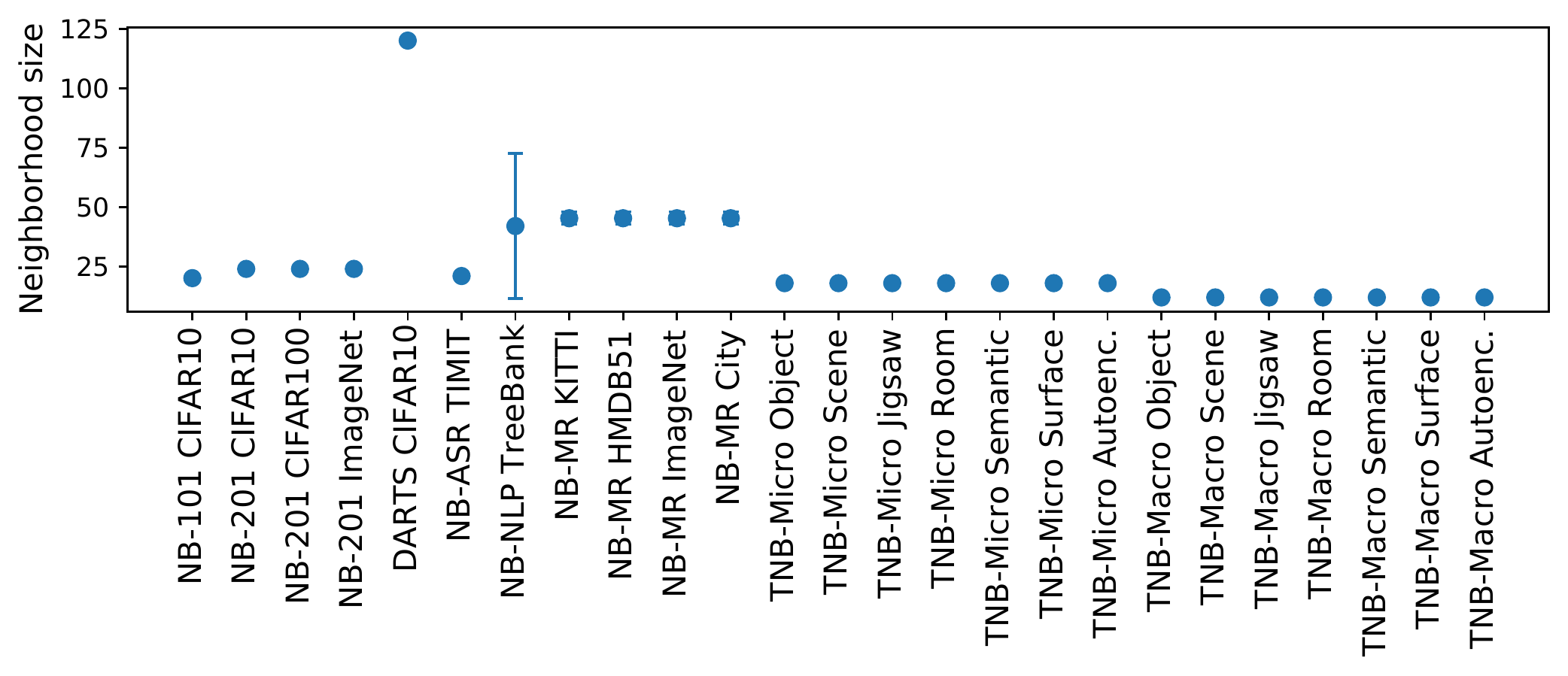}
    \caption{Average neighborhood size for each NAS benchmark. Note that for some NAS benchmarks,
    the neighborhood size is fixed, and for other NAS benchmarks, the neighborhood size varies.
    }
    \label{fig:nbhd_sizes}
\end{figure}

We start by giving additional experiments on the statistics of NAS benchmarks, to supplement Section \ref{sec:statistics}.
In Figure \ref{fig:box_plots_appendix}, we give the full box plots, extending Figure \ref{fig:box_plots}.
We also add new statistics:
in Figure \ref{fig:runtimes}, we plot the average time to train an architecture for each NAS benchmark. 
In Figure \ref{fig:nbhd_sizes}, we plot the average neighborhood size for each NAS benchmark. Note that for some NAS benchmarks, the neighborhood size is fixed, and for other NAS benchmarks, the neighborhood size varies.

\subsection{Additional Algorithm Experiments} \label{app:algo_experiments}

\begin{figure}[t]
    \centering
    \includegraphics[width=.99\textwidth]{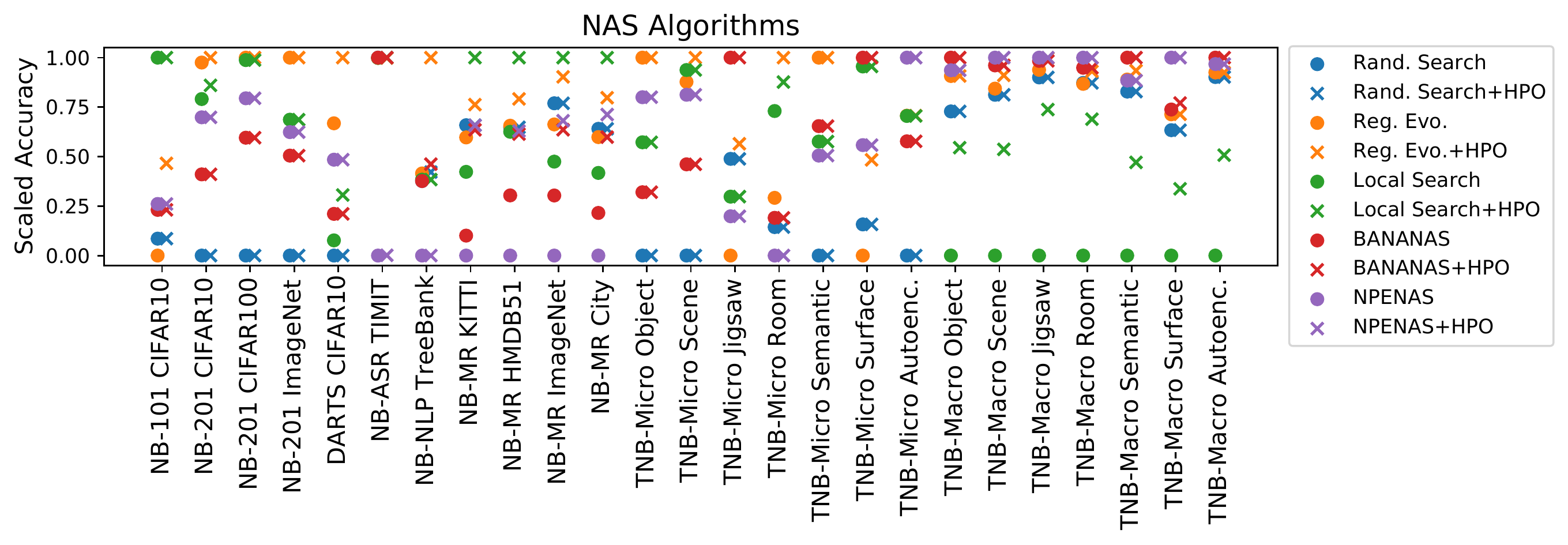}
    \includegraphics[width=.99\textwidth]{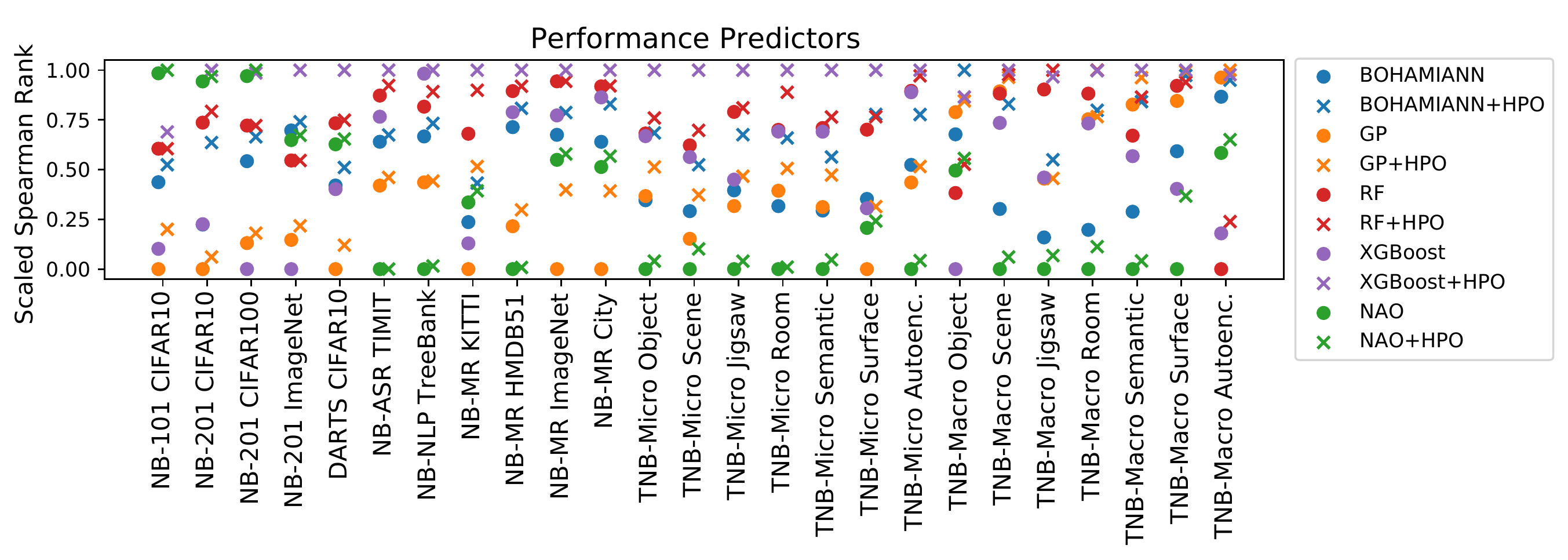}
    \caption{
    Relative performance of black-box algorithms (top) and performance predictors (bottom) across NAS benchmarks. The solid circle shows the performance of the algorithm with default hyperparameters, while the cross shows performance after hyperparameter optimization.
    }
    \label{fig:predictors_appendix}
\end{figure}

Next, we give additional experiments from Section \ref{sec:experiments}.
In Figure \ref{fig:predictors_appendix}, we give the full performance predictor and black-box results, extending Figure \ref{fig:predictors}.

Recall that in Section \ref{sec:experiments}, we assessed the
transferability of hyperparameters by tuning algorithms on NAS benchmark $A$, 
and evaluating the performance of the tuned method on NAS benchmark $B$, compared to the performance of the best hyperparameters from NAS benchmark $B$. 
The results were plotted in Figure \ref{fig:matrices}. 
In Tables \ref{tab:pred_matrix_raw} and \ref{tab:bbo_matrix_raw}, to present the results in another format, we give the raw values from these experiments. All values are averaged over each search space. For example, the three rows for NAS-Bench-201 were averaged into one row, and similarly for the columns.

Furthermore, in Tables \ref{tab:pred_matrix_summary} and \ref{tab:bbo_matrix_summary}, we give summary statistics from this experiment. For each search space, we compute the transferability of hyperparameters on average to or from all other search spaces. 
For performance predictors, we find that hyperparameters from NAS-Bench-MR transfer the least well to or from other search spaces. NAS-Bench-NLP transfers the best.
For black-box algorithms, NAS-Bench-MR transfers the worst, and DARTS transfers the best.
Therefore, it is safest for practitioners to tune their techniques on NAS-Bench-NLP and DARTS before deploying them in a new setting. It is not safe to tune techniques on other benchmarks such as NAS-Bench-MR.

While all of the hyperparameter transfer experiments to this point have focused on the optimal hyperparameters, we now present two more experiments that focus on the hyperparameters on average.
In Figure \ref{fig:corr_matrices}, for search spaces A and B, we compute the Kendall Tau rank correlation of the ranking lists for \emph{all} hyperparameters on search spaces A and B.

Finally, in Table \ref{tab:loo}, we present ``leave one out'' experiments for all search spaces.
For a search space A, the best hyperparameter setting on average over all search spaces except A is computed, and compared to the performance of the best hyperparameters for A. 
This is similar to Table \ref{tab:pred_matrix_summary} in that it provides a summary of the average transferability of the hyperparameters for each search space. However, the leave one out experiments are focused more on the performance of hyperparameters on average, while  Table \ref{tab:pred_matrix_summary} is focused on transferability of the optimal hyperparameters.

\subsection{A guide to interpreting hyperparameter transfer experiments} \label{app:hpo_guide}
Throughout Sections \ref{sec:experiments} and \ref{app:algo_experiments}, we presented several different analyses and summaries on the extent to which hyperparameters transfer from one search space to another. In this section, we give a guide for interpreting the results.

\begin{figure}[t]
    \centering
    \includegraphics[width=.493\textwidth]{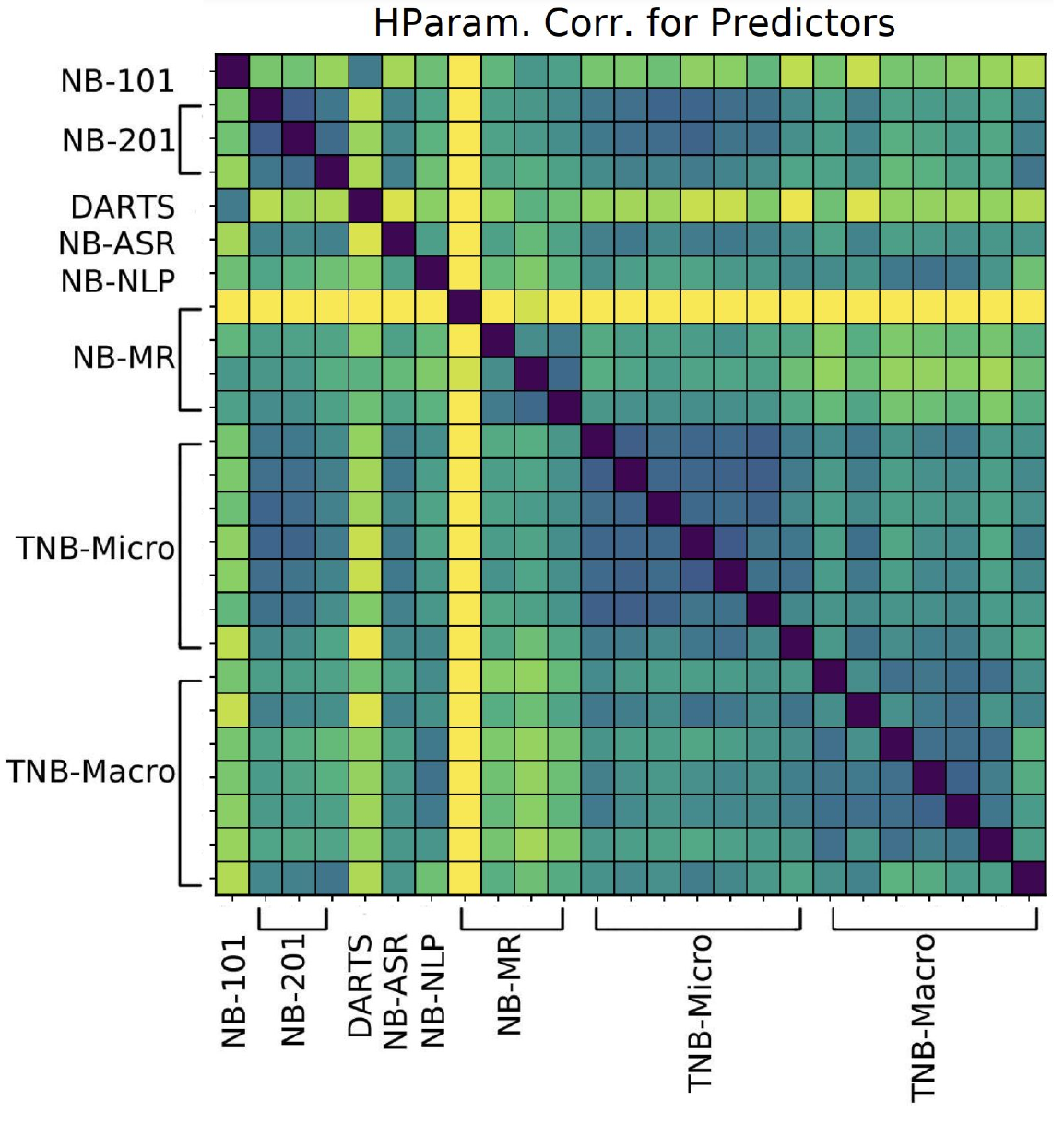}
    \includegraphics[width=.483\textwidth]{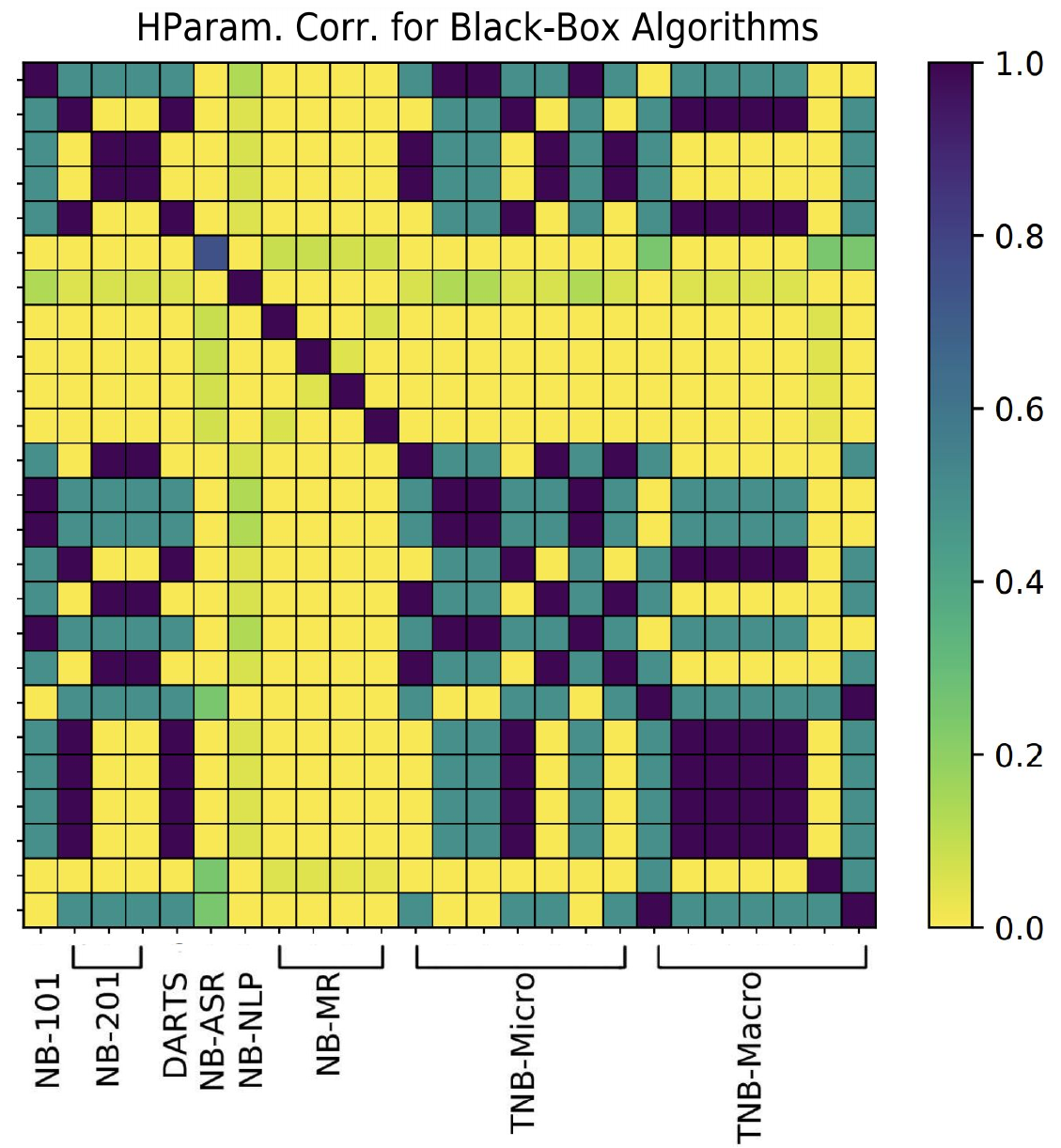}
    \caption{Transferability results for predictors (left) and black-box algorithms (right). 
    Row $i$, column $j$ denotes the Kendall Tau rank correlation of the performance of hyperparameters between search spaces $i$ and $j$. 
    For abbreviations, see Table \ref{tab:abbreviations}.
    }
    \label{fig:corr_matrices}
    \vspace*{-0.1cm}
\end{figure}

First, practitioners interested in the transfer of hyperparameters which are \emph{optimally trained} on one search space, should focus on Figure \ref{fig:matrices} and Tables \ref{tab:pred_matrix_summary} and \ref{tab:bbo_matrix_summary}, because these figures and tables express the regret of hyperparameters tuned on one search space and evaluated on others.
On the other hand, practitioners interested on how hyperparameters overall transfer from one search space to others (not just optimal), should focus on Figure \ref{fig:corr_matrices} and Table \ref{tab:loo}, because these represent the average transferability of all sets of hyperparameters that we tried.

Practitioners interested in the specific transfer from one search space (or setting) to another, should focus on our matrix results, Figures \ref{fig:matrices} and \ref{fig:corr_matrices}. Practitioners interested in the general transferability on average to or from one search space to others, should focus on the summary tables, Tables \ref{tab:pred_matrix_summary}, \ref{tab:bbo_matrix_summary}, and \ref{tab:loo}.

\begin{table}[t]
\caption{Raw values from the performance predictor hyperparameter transferability experiment from Figure \ref{fig:matrices} (left).
Each search space has 0-1 scaling done to fairly compare trends between search spaces.
Results are weighted by search space. E.g., each of the three NAS-Bench-201 benchmarks are averaged into one row/column.}
\centering
\begin{tabular}{@{}l|c|c|c|c|c|c|c@{}}
\toprule
& \multicolumn{1}{c}{\textbf{NB-101}} & \multicolumn{1}{c}{\textbf{NB-201}} & \multicolumn{1}{c}{\textbf{DARTS}} & \multicolumn{1}{c}{\textbf{NB-ASR}} & 
\multicolumn{1}{c}{\textbf{NB-NLP}} & \multicolumn{1}{c}{\textbf{NB-MR}} & 
\multicolumn{1}{c}{\textbf{TNB-101}} \\
\midrule 
\textbf{NB-101} & .00 & .42 & .28 & .46 & .48 & .21 & .41 \\
\textbf{NB-201} & .43 & .02 & .29 & .11 & .11 & .32 & .09 \\
\textbf{DARTS} & .08 & .47 & .00 & .48 & .34 & .26 & .41 \\
\textbf{NB-ASR} & .28 & .23 & .31 & .00 & .07 & .32 & .12 \\
\textbf{NB-NLP} & .25 & .31 & .27 & .09 & .00 & .36 & .15 \\
\textbf{NB-MR} & .19 & .39 & .28 & .42 & .47 & .20 & .40 \\
\textbf{TNB-101} & .37 & .13 & .41 & .16 & .15 & .40 & .15 \\
\bottomrule
\end{tabular}
\label{tab:pred_matrix_raw}
\end{table}

\begin{table}[t]
\caption{Raw values from the black-box algorithm hyperparameter transferability experiment from Figure \ref{fig:matrices} (right).
Each search space has 0-1 scaling done to fairly compare trends between search spaces.
Results are weighted by search space. E.g., each of the three NAS-Bench-201 benchmarks are averaged into one row/column.}
\centering
\begin{tabular}{@{}l|c|c|c|c|c|c|c@{}}
\toprule
& \multicolumn{1}{c}{\textbf{NB-101}} & \multicolumn{1}{c}{\textbf{NB-201}} & \multicolumn{1}{c}{\textbf{DARTS}} & \multicolumn{1}{c}{\textbf{NB-ASR}} & 
\multicolumn{1}{c}{\textbf{NB-NLP}} & \multicolumn{1}{c}{\textbf{NB-MR}} & 
\multicolumn{1}{c}{\textbf{TNB-101}} \\
\midrule 
\textbf{NB-101} & .00 & .25 & .25 & .00 & .25 & .75 & .27 \\
\textbf{NB-201} & .25 & .22 & .33 & .75 & .50 & .83 & .28 \\
\textbf{DARTS} & .25 & .33 & .00 & .75 & .50 & .50 & .23 \\
\textbf{NB-ASR} & .75 & .50 & .50 & .00 & .75 & .25 & .50 \\
\textbf{NB-NLP} & .07 & .28 & .28 & .73 & .00 & .64 & .28 \\
\textbf{NB-MR} & .86 & .80 & .70 & .42 & .84 & .32 & .76 \\
\textbf{TNB-101} & .27 & .28 & .23 & .73 & .48 & .70 & .28 \\
\bottomrule
\end{tabular}
\label{tab:bbo_matrix_raw}
\end{table}

\begin{table}[t]
\caption{Summaries from the performance predictor transferability experiment from Figure \ref{fig:matrices} (left).
Each value in ``transfer to'' is the average of the corresponding row from Figure \ref{fig:matrices}. Each value in ``transfer from'' is the average of the corresponding column.
Therefore, for each search space, we have a measure of the extent to which hyperparameters can transfer to or from other search spaces, on average.
}
\begin{tabular}{@{}l|c|c|c|c|c|c|c@{}}
\toprule
& \multicolumn{1}{l}{\textbf{NB-101}} & \multicolumn{1}{c}{\textbf{NB-201}} & \multicolumn{1}{c}{\textbf{DARTS}} & \multicolumn{1}{c}{\textbf{NB-ASR}} & 
\multicolumn{1}{c}{\textbf{NB-NLP}} & \multicolumn{1}{c}{\textbf{NB-MR}} & 
\multicolumn{1}{c}{\textbf{TNB-101}} \\
\midrule 
Transfer to & 0.376 & 0.229 & 0.340 & 0.224 & 0.239 & 0.393 & 0.293 \\
Transfer from & 0.268 & 0.328 & 0.307 & 0.288 & 0.270 & 0.345 & 0.287 \\
\bottomrule
\end{tabular}
\label{tab:pred_matrix_summary}
\end{table}

\begin{table}[t]
\caption{Summaries from the black-box algorithm transferability experiment from Figure \ref{fig:matrices} (right).
Each value in ``transfer to'' is the average of the corresponding row from Figure \ref{fig:matrices}. Each value in ``transfer from'' is the average of the corresponding column.
Therefore, for each search space, we have a measure of the extent to which hyperparameters can transfer to or from other search spaces, on average.
}
\begin{tabular}{@{}l|c|c|c|c|c|c|c@{}}
\toprule
& \multicolumn{1}{l}{\textbf{NB-101}} & \multicolumn{1}{c}{\textbf{NB-201}} & \multicolumn{1}{c}{\textbf{DARTS}} & \multicolumn{1}{c}{\textbf{NB-ASR}} & 
\multicolumn{1}{c}{\textbf{NB-NLP}} & \multicolumn{1}{c}{\textbf{NB-MR}} & 
\multicolumn{1}{c}{\textbf{TNB-101}} \\
\midrule 
Transfer to & 0.461 & 0.528 & 0.428 & 0.542 & 0.381 & 0.784 & 0.495 \\
Transfer from & 0.407 & 0.444 & 0.383 & 0.731 & 0.555 & 0.666 & 0.433 \\
\bottomrule
\end{tabular}
\label{tab:bbo_matrix_summary}
\end{table}

\begin{table}[t]
\caption{Leave one out experiments for performance predictors.
For a search space A, the best hyperparameter setting on average over all search spaces except A is computed, and compared against the best hyperparameter setting of search space A when transferring to (or from) search space A. 
}
\begin{tabular}{@{}l|c|c|c|c|c|c|c@{}}
\toprule
& \multicolumn{1}{l}{\textbf{NB-101}} & \multicolumn{1}{c}{\textbf{NB-201}} & \multicolumn{1}{c}{\textbf{DARTS}} & \multicolumn{1}{c}{\textbf{NB-ASR}} & 
\multicolumn{1}{c}{\textbf{NB-NLP}} & \multicolumn{1}{c}{\textbf{NB-MR}} & 
\multicolumn{1}{c}{\textbf{TNB-101}} \\
\midrule 
Transfer to & 0.372 & 0.37 & 0.48 & 0.195 & 0.185 & 0.445 & 0.252 \\
Transfer from & 0.226 & 0.16 & 0.223 & 0.058 & 0.029 & 0.241 & 0.056 \\
\bottomrule
\end{tabular}
\label{tab:loo}
\end{table}


\subsection{Insight Experiments} \label{app:insight_experiments}

Now we present experiments that give new insights into NAS algorithms and search spaces.

\begin{figure}[t]
    \centering
    \includegraphics[width=.4\textwidth]{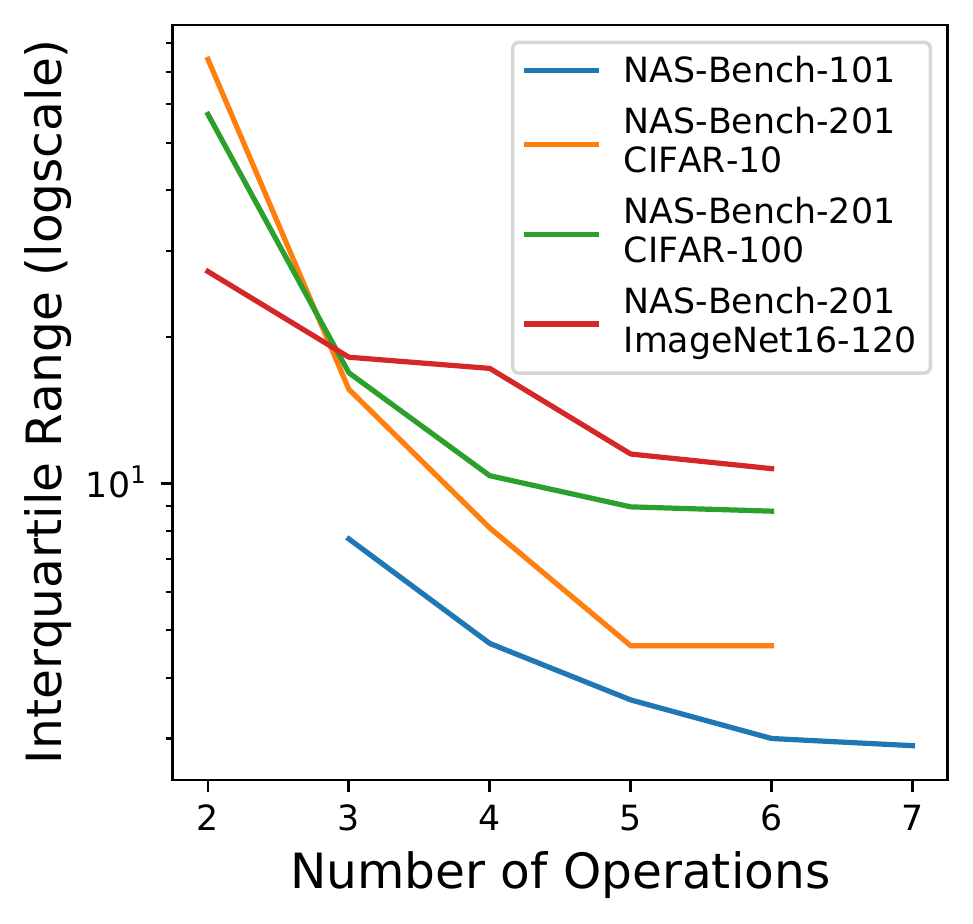}
    \caption{Interquartile ranges of subsets of NAS-Bench-101 and NAS-Bench-201
    as a function of the number of operations. As the size of the search space increases,
    the interquartile range decreases.}
    \label{fig:iqr}
\end{figure}

First, we run experiments to test the following hypothesis: larger search spaces have smaller interquartile ranges (IQR), because a larger cell size gives most architectures a chance to have performant operations. For example, for a search space of size three, some architectures will have two or three convolution operations, and some architectures will have two or three pooling operations, creating a large IQR. But for a search space of size 10, the vast majority of architectures will have a good mix of convolution and pooling operations, creating a comparatively lower IQR. 
We run this experiment on NAS-Bench-101 and NAS-Bench-201.
See Figure \ref{fig:iqr}.
We find that in all benchmarks, there is a strict negative correlation between number of operations and IQR.

Finally, we compute correlations between the relative ranking of each NAS technique and properties of the search spaces, such as total size and neighborhood size.
See Figure \ref{tab:correlation_insights}.
The largest correlations we find are as follows: 
\begin{itemize}
    \item GP performs comparatively much better on search spaces with small neighborhood sizes.
    \item When tuned, RF and XGBoost perform comparatively much better on large search spaces and also search spaces with large neighborhood sizes.
    \item Surprisingly, BOHAMIANN performs comparatively better for large neighborhood sizes when not tuned, and comparatively better for small neighborhood sizes when tuned.
    \item Default regularized evolution performs comparatively much better on search spaces with small neighborhood sizes.
    \item Local search performs comparatively better on small search spaces and also search spaces with small neighborhood sizes.
\end{itemize}

\begin{table}[h]
\caption{Correlation insights for five NAS algorithms (left) or five performance predictors (right).
Kendall Tau rank correlations are computed between properties of the search spaces (search space size or neighborhood size) and the relative ranking list for predictors or NAS algorithms, with or without HPO.
Since a ranking list is used, a \emph{negative} correlation means a \emph{positive} correlation between the search space property and algorithmic performance.}
    \label{tab:correlation_insights}
\centering
    \resizebox{\textwidth}{!}{
\begin{tabular}{@{}lcccccccccc@{}}
\toprule
\textbf{} & \multicolumn{5}{c}{\textbf{Performance Predictors}} & \multicolumn{5}{c}{\textbf{NAS Algorithms}} \\
&  BOHAM. & GP & RF & XGB & NAO & RS & RE & BANANAS & LS & NPENAS \\ 
\midrule
Default vs.\ SS size & -.07 & .25 & -.16 & -.39 & \multicolumn{1}{c|}{.20} &
-.29 & -.21 & -.15 & .48 & .26 \\
HPO vs.\ SS size & .15 & .05 & -.51 & -.39 & \multicolumn{1}{c|}{.20} & 
-.26 & -.21 & -.05 & .21 & .35 \\
Default vs.\ Nbhd.\ size & -.33 & .45 & -.26 & .00 & \multicolumn{1}{c|}{.00} & 
-.10 & -.51 & .25 & .37 & .16 \\
HPO vs.\ Nbhd.\ size & .35 & .37 & -.39 & -.51 & \multicolumn{1}{c|}{.00} & 
-.16 & -.31 & .05 & .31 & .05 \\
\bottomrule
\end{tabular}
}
\end{table}

%% file: sections/C_appendix_codebase.tex
\section{Details from Section~\ref{sec:codebase}} 
\label{app:codebase}

In this section, we give more details on the API for \nasbs.

When designing \nasbs{}, we strove for both minimalism and generalism. In order to add a new benchmark in \nasbs{}, one has to first define the computational graph using the NetworkX API. This graph encompasses high-level abstractions such as the $\texttt{add\_node}$ and $\texttt{add\_edges\_from}$ methods. If we are implementing a tabular or surrogate benchmark as the ones used in Section \ref{sec:experiments}, a $\texttt{get\_dataset\_api}$ function needs to be implemented, which is used as an interface to the original pre-computed benchmark data. The PyTorch computational graph is generated via the $\texttt{adapt\_search\_space}$ method of the NAS algorithms or performance predictors. For instance, it can determine if operation choices in edges should be combined as a mixed operation (as done in DARTS~\citep{darts}) or if they should be categorical choices from which the NAS algorithms sample. Afterwards, the graph instance is stored as an attribute of the optimizer instance. The $\texttt{Trainer}$, which runs the optimization loop, interacts only with the optimizer (see line 16 in Snippet~\ref{fig:one-line}).